\documentclass[12pt,onecolumn,twoside]{IEEEtran}
\usepackage{epsfig}
\usepackage{psfig}
\usepackage{floatflt} % text flows around figures
\usepackage{paralist} % compact and in-paragraph lists
\usepackage{latexsym}
\usepackage{amsfonts}
\usepackage{amssymb}
\usepackage{amstext}

\newcommand{\bE}{\mathbf{E}}

\newcommand{\bH}{\mathbf{H}}
\newcommand{\bI}{\mathbf{I}}

\newcommand{\bP}{\mathbf{P}}

\newcommand{\bR}{\mathbf{R}}

\newcommand{\bW}{\mathbf{W}}
\newcommand{\bGamma}{\mathbf{\Gamma}}

\newcommand{\bTheta}{\mathbf{\Theta}}

\newcommand{\bzero}{\mathbf{0}}

\newcommand{\bm}{\mathbf{m}}

\newcommand{\bx}{\mathbf{x}}

\newcommand{\bgamma}{\mbox{\boldmath{$\gamma$}}}

\newcommand{\bdelta}{\mbox{\boldmath{$\delta$}}}
\newcommand{\btheta}{\mbox{\boldmath{$\theta$}}}
\newcommand{\bnabla}{\mbox{\boldmath{$\nabla$}}}

\newcommand{\cR}{\mathcal{R}}

\newcommand{\R}{{\mathbb{R}}}

\begin{document}

\title{Maximum Likelihood Mosaics}

\author{Bernardo Esteves Pires, \ Pedro M.~Q.~Aguiar,~{\sl Member,~IEEE}% <-this % stops a space
\thanks{Contact author: P. Aguiar, ISR---Institute for Systems and Robotics, Instituto
Superior T\'{e}cnico, Av. Rovisco Pais, 1049-001 Lisboa, Portugal.
E-mail: {\tt aguiar@isr.ist.utl.pt}. His work was partially
supported by FCT
grant POSI/SRI/41561/2001.}% <-this % stops a space
\thanks{B. Pires is
with The Boston Consulting Group, Lisbon Office, Portugal. E-mail:
{\tt pires.bernardo@bcg.com}.}}

%\markboth{IEEE Transactions on Image Processing}{Pires and Aguiar:
%Maximum Likelihood Mosaics}

\maketitle

\begin{abstract}
The majority of the approaches to the automatic recovery of a
panoramic image from a set of partial views are suboptimal in the
sense that the input images are aligned, or registered, pair by
pair, {\it e.g.}, consecutive frames of a video clip. These
approaches lead to propagation errors that may be very severe,
particularly when dealing with videos that show the same region at
disjoint time intervals. Although some authors have proposed a
post-processing step to reduce the registration errors in these
situations, there have not been attempts to compute the optimal
solution, {\it i.e.}, the registrations leading to the {\it
panorama that best matches the entire set of partial views}. This
is our goal. In this paper, we use a generative model for the
partial views of the panorama and develop an algorithm to compute
in an efficient way the Maximum Likelihood estimate of all the
unknowns involved: the parameters describing the alignment of all
the images and the panorama itself.
\end{abstract}


%\begin{center}{\bfseries EDICS:
%2-ANAL} (Image and Video Processing---Analysis), {\bfseries
%2-SEQP} (Image Sequence Processing), {\bfseries 2-INTR}
%(Interpolation and Spatial Transformations).\end{center}

\begin{center} Permission to publish this abstract separately
is granted. \end{center}

%\newpage

\section{Introduction}
\label{sec:intro}

Very good paper to read. Technically sound, very clear, and very
good results. My only minor comment is that the statement that no
attempt has been made to find the optimal solution is not true.
There are many papers tackling this issue, e.g. work of Pollefeys,
zisserman's group at oxford, some work at INRIA by various groups,
the work of Peleg et al., etc.

In this paper, we address the problem of recovering, in an
automatic way, a panoramic image, or a mosaic, from a set of
uncalibrated partial views, {\it e.g.}, a set of video frames.
Modern digital video systems demand efficient solutions for this
problem, {\it e.g.}, for image
stabilization~\cite{dufaux00,petrovic04} and content-based
representations~\cite{aguiarjasinschimourapluem04}. Other
application fields include virtual reality and remote sensing. The
key step to the success of the automatic mosaic building is the
accurate registration, or alignment, of the input images.

\subsection{Related work}

Although some authors have approached the registration problem
using classical signal processing techniques, such as Fourier
transforms~\cite{reddy96}, or current image analysis tools, such
as integral projections~\cite{lee02}, the majority of the papers
in the literature are mostly distinguished by either requiring a
low-level pre-processing step (feature-based methods) or
attempting to register the images directly from their intensity
levels (featureless methods).

Feature-based methods, {\it e.g.}, \cite{hartley2000}, align the
 images by first detecting and matching a set of pointwise
features. Since reliable feature points must correspond to sharp
intensity corners~\cite{shi94,aguiar01-icip}, this first step is
hard to accomplish in a fully automatic way when processing real
videos, particularly when the images are noisy, have low texture,
or exhibit a small overlap among them.

In opposition, featureless methods are optimal, in the sense that
they estimate the registration parameters by minimizing the
difference between the image intensities in a large region, thus
leading to more robust solutions to the registration of a pair of
views, {\it e.g.}, \cite{mann97,pires04}. However, when building a
panorama from a large set of images, practitioners usually
register them sequentially, one at a time. This leads to
propagation errors that may be become visually noticeable if
non-consecutive images cover the same region of the panorama,
which is common in applications such as seabed mapping. Although
some authors proposed to post-process the registration parameters
to deal with this problem~\cite{mann97,hasler99}, there have not
been attempts to generalize the highly successful featureless
methods to the multi-frame case.

\subsection{Proposed approach: featureless global estimation}

The robustness of the featureless approaches to the registration
of two views motivated us to develop a featureless method to align
a larger set of frames. However, it is not obvious how the
two-frame cost function, usually the sum of the image square
differences~\cite{mann97,pires04}, should be generalized to the
multi-frame case. We were able to derive the appropriate cost
function, which is an original contribution of this paper, by
including as unknown, jointly with the registration parameters,
the panoramic image itself.

Our approach in this paper is then to formulate the automatic
recovery of mosaics from a set partial views, as a classical
parameter estimation problem. The input images are modelled as
noisy observations of limited regions of the unknown panorama.
Naturally, since the images are uncalibrated, the problem includes
as unknowns the parameters describing the registration, or
alignment, of the entire set of input images. We then use {\it
Maximum Likelihood}~(ML) estimation. To minimize the ML~cost with
respect to the large set of unknowns, we propose an efficient
method. First, we derive the closed-form solution for the estimate
of the panorama in terms of the other unknowns (the registration
parameters). Then, we plug-in the estimate of the panorama into
the ML~cost, obtaining an error function that depends on the
registration parameters alone. This error function is a weighted
sum of the square differences between all possible pairs of input
images. We derive a gradient-descent algorithm to minimize this
cost.

Like in the current featureless approaches to the registration of
two images~\cite{mann97,pires04}, the derivatives involved in the
gradient-descent algorithm to minimize our ML cost, are computed
in a simple way in terms of the image gradients.

\subsection{Paper organization}

The remaining of the paper is organized as follows. In
section~\ref{sec:formulation}, we formulate the registration of
multiple images as a classical estimation problem.
Section~\ref{sec:mlm} deals with ML estimation for this problem,
{\it i.e.}, it introduces the {\it Maximum Likelihood
Mosaics}~(MLM) approach. In section~\ref{sec:tfr}, we develop MLM,
using the simpler case of registering a pair of images. We
contrast MLM with minimizing the registration error over a fixed
window, as usually done in current featureless approaches.
Section~\ref{sec:mfr} generalizes MLM to the multi-frame
registration. In section~\ref{sec:mlma}, we derive the
gradient-descent algorithm to minimize the ML cost.
Section~\ref{sec:exp} describes experiments and
section~\ref{sec:conc} concludes the paper. Preliminary versions
of parts of this work are in~\cite{pires04,pires05}.

\section{Problem Formulation}
\label{sec:formulation}

In this section, we develop a generative model for the partial
views of an unknown panorama, and use ML to derive the estimation
criterion that will allow us to recover the observed panorama, as
well as the registration parameters, {\it i.e.}, the viewing
positions.

\subsection{Generative model}

We model each pixel of each image $\bI_i$, as a noisy sample of
the panorama $\bP$. For simplicity, we consider the image domain
to be the entire plane~$\R^2$ and, to take care of the limited
field of view, we define a window~$\bH$ as $\bH(x,y)\!=\!1$ in the
region observed in the images and~$\bH(x,y)\!=\!0$ in the regions
outside the camera field of view. The observation model is then
\begin{equation}
\bI_i(\bx_i)=\bigl[\bP(\bx_0)+\bR\left(\bx_i\right)\bigr]\bH(\bx_i)\,,\label{eq:obs}
\end{equation}
% . Using $\bI^i$
%to refer to a generic picture in the set we can write:
%$\bI^i(\bx^i)=\bP(\bx^0)+\bR(\bx^i)$
where $\bR$ denotes the noise, assumed i.i.d.~zero-mean Gaussian,
$\bx_i$~are the image coordinates~$(x,y)$, expressed in the
coordinate system of the generic image $\bI_i$, and $\bx_0$~are
the corresponding coordinates of the panoramic image $\bP$,
expressed in its own coordinate system (which we will refer to as
the reference coordinate system). Image models related
to~(\ref{eq:obs}) have been used in the context of segmenting and
tracking moving objects in video
sequences~\cite{aguiar97,jojic01}.

The reference coordinate system and the coordinate system of any
of the images are related by a generic parametric mapping
\begin{equation}
\bx_i=\bm(\btheta_i;\bx_0)\,.\label{eq:map}
\end{equation}
The parameter vector~$\btheta_i$ in~(\ref{eq:map}) determines thus
the mapping between each pixel of the panorama, with
coordinates~$\bx_0$, expressed in the reference coordinate system,
with the corresponding pixel of image~$\bI_i$, with
coordinates~$\bx_i$. Common parameterizations include translation
(2~degrees of freedom (dof)), rotation (1~dof), rigid motion
(3~dof), translation+rotation+zoom (4~dof), affine (6~dof), and
the projective, or homography (8~dof), see, {\it e.g.},
\cite{mann97,kim03}. Although our derivations are intentionally
left fully generic, in the experiments, we have used the affine
mapping.

\subsection{Estimation criterion}

Given a set of $n$ images, $\left\{\bI_1,\ldots,\bI_n\right\}$,
our goal is to recover all the unknowns involved: the
panorama~$\bP$ and the set of parameter vectors
$\left\{\btheta_1,\ldots,\btheta_n\right\}$ that define the
viewing positions. We use~ML.
%Using the parametric
%mapping we've introduced and taking the usual assumptions that the
%noise in each pixel is normal with zero mean and $\sigma^2$
%variance and is independent from pixel to pixel and from image to
%image we can write
From the observation model~(\ref{eq:obs}), after simple
manipulations, we express the symmetric of the log-likelihood
function~as
\begin{equation}
L\left(\bP,\btheta_1,\ldots,\btheta_n\right)=
\frac{nN}{2}\ln\left(2\pi\sigma^{2}\right)
 +  \left(2\sigma^{2}\right)^{-1}
 \label{eq:1}
\sum_{\bx_0\in\R^2}\sum_{i=1}^n\bigl[\bI_i\left(\bm(\btheta_i;\bx_0)\right)
\!-\!\bP\left(\bx_0\right)\bigr]^2
\,\bH\left(\bm(\btheta_i;\bx_0)\right), \nonumber\end{equation}
where $N$ is the number of pixels in each image and $\sigma^2$ is
the variance of the observation noise.
%
%$\bh^i$ is a window function equal to $1$ in the region of $\bx^0$
%where the generic image $\bI^i$ was sampled (assuming the
%parameters of the image mapping are $\btheta^i$) and equal to $0$
%elsewhere.

\section{Maximum Likelihood Mosaics}
\label{sec:mlm}

To compute the ML estimate of all the unknowns, {\it i.e.}, to
carry out the minimization of the ML~cost, given by the symmetric
log-likelihood~(\ref{eq:1}), with respect to (wrt)
$\left\{\bP,\btheta_1,\ldots,\btheta_n\right\}$, we start by
noticing that the estimate of the panorama~$\bP$ can be expressed
in closed-form as a function of the remaining unknowns.

%\subsection{Estimate of the panorama~$\bP$}

We derive the expression for the ML estimate~$\widehat{\bP}$ of
the panorama by minimizing~(\ref{eq:1}) wrt a generic pixel
value~$\bP(\bx_0)$. By making zero the derivative of~(\ref{eq:1})
wrt~$\bP(\bx_0)$, the estimate~$\widehat{\bP}$ at pixel~$\bx_0$ is
easily obtained as a function of the set of unknown registration
parameters, which we will compactly denote by
$\bTheta=\left\{\btheta_1,\ldots,\btheta_n\right\}$:
\begin{equation}
\widehat{\bP}\left(\bx_0,\bTheta\right)= \frac{ \sum_{i=1}^n
\bI_i\left(\bm(\btheta_i;\bx_0)\right)\,
\bH\left(\bm(\btheta_i;\bx_0)\right)} { \sum_{i=1}^n
\bH\left(\bm(\btheta_i;\bx_0)\right)}\,.
\label{eq:2}\end{equation} This expression shows that the estimate
of the intensity of each pixel $\bx_0$ of $\widehat{\bP}$ is given
by the average of the intensities of the corresponding pixels of
all the input images that captured $\bx_0$, \textit{i.e.}, all the
images~$\bI_i$ for which
$\bH\left(\bm(\btheta_i;\bx_0)\right)\!=\!1$.

\section{Two-frame Registration}
\label{sec:tfr}

\subsection{Cost-function}

\subsection{Minimization algorithm}

It is now clear that it is not possible to evaluate the error
$e(\btheta,\bx)$ in~(\ref{eq:funcional}) for every pair of values
of $\btheta$ and $\bx$.
%, in particular it is
%not possible to know the value of the registration
%error~$E(\btheta)$ error for values of $\btheta$ such that there
%is no overlap between the images.
This is the main problem of using a fixed
window~$\cR$---registration is only possible when the overlapping
region contains $\cR$.
This limitation has a  particular impact on the behavior of
iterative registration algorithms. In fact, to avoid an exhaustive
search, {\it e.g.}, block matching, the minimization
of~$E(\btheta)$ in~(\ref{eq:funcional}) is usually performed by
using gradient-based algorithms that iteratively
optimize~$\btheta$.
% The
%initialization is taken care of by using a multiresolution scheme.
Obviously, at every iteration of the algorithm, the overlapping
region (which depends on the current estimate of $\btheta$) must
contain~$\cR$, since the error $E(\btheta)$, as well as its
gradient, depend on a sum over~$\cR$.

As illustrated by the first experiment of section~\ref{sec:exp},
the minimum overlap requirement makes hard the automatic
registration of arbitrary images. In fact, by specifying {\it a
priori} a fixed window $\cR$, we can not cope with all possible
situations. If one specifies a small $\cR$, it may fit into the
true overlapping region, but the estimation error will be large
due to the smooth minimum of $E(\btheta)$. On the other hand, if
one specifies a large $\cR$, it is not possible to register images
which have a small overlap. Our goal here is to develop a method
to perform the registration in the situations when the overlap
between the images is not known \textit{a priori}.

Instead of using a fixed window $\cR$, we propose an adaptive
window $\cR_A(\btheta)$, defined as the {\it largest region} for
which it is possible to evaluate the error $e(\btheta,\bx)$ as
defined in~(\ref{eq:funcional}). In our iterative optimization,
the estimate $\btheta$ is computed by refining a previous estimate
$\btheta_0$, {\it i.e.}, $\btheta=\btheta_0+\bdelta$. The
update~$\bdelta$ is estimated by minimizing the registration error
over the adaptive window $\cR_A(\btheta_0)$,
\begin{equation}
\widehat{\bdelta}=\arg\min_{\bdelta}\sum_{\bx \in
\cR_A(\btheta_0)} e^2(\btheta_0+\bdelta,\bx) ,
\label{eq:funcional2}
\end{equation}
where $e(\btheta,\bx)$ is as defined in~(\ref{eq:funcional}). The
adaptive window $\cR_A(\btheta_0)$, whose size and shape depend on
the current estimate~$\btheta_0$ of the motion parameter vector,
is the overlapping region between the image~$\bI$ and the
image~$\bI'$ registered according to $\btheta_0$,
$\;\bI'(\bm(\btheta_0,\bx))$.

To compute the update $\bdelta$, we develop an
adaptive-window-based Gauss-Newton method. Similar methods have
been used to minimize~(\ref{eq:funcional}), {\it i.e.}, to
register images using a fixed window, {\it e.g.}, \cite{mann97}.
In this method, $e(\btheta,\bx)$ is approximated by its
first-order truncated Taylor series expansion, $e(\btheta,\bx)
\simeq e(\btheta_0,\bx)+\bdelta^T \cdot
\nabla_{\btheta}e(\btheta_0,\bx)$. Using this approximation
in~(\ref{eq:funcional2}), and making zero the gradient of the cost
function, we get $\widehat\bdelta$ as the solution of the linear
system
\begin{equation}\left(\sum_{\bx \in
\cR_A(\btheta_0)} \!\!\bnabla_{\btheta} e
\cdot\!\bnabla_{\btheta}^T e\right) \,\cdot \,\, \widehat\bdelta
\,+\, \sum_{\bx \in \cR_A(\btheta_0)} \!\!e\,\bnabla_{\btheta} e\,
=\, \bzero \,,\label{eq:pa}
\end{equation}
where we omit the dependency of $e$ on $\bx$ and $\btheta_0$ for
compactness. From the definition of~$e$ in~(\ref{eq:funcional}),
we see that $\bnabla_{\btheta} e$ in~(\ref{eq:pa}) is computed
from the image gradient, $\bnabla_{\btheta}e
 = -{\bnabla_{\btheta}}\bm\cdot{\bnabla_{\bx}}\bI'$.
%, see
%also~\cite{mann97}.
%\bI'\!\cdot\!{\bnabla_{\bx}} \bI'$
%
%However, the whole process assumes that the estimator offsets are
%small. This can very easily be untrue so,
%
The initial guess for $\btheta_0$ is such that
$\bm(\btheta_0,\bx)$ is the identity mapping, which corresponds to
initializing the algorithm with zero displacement between the
images, thus the initial window $\cR_A(\btheta_0)$ is the entire
image region.

The Gauss-Newton method just described assumes the motion is
small. To cope with large displacements, we use a multiresolution
scheme. In such scheme, the iterative estimation algorithm is
first used in a lower resolution versions of the input images,
until a certain stopping criterium is reached . The resulting
parameter estimates are then used as initial guesses for the
parameters in the the next (higher) resolution and the process is
repeated until the original images are used.

Among the number of valid stopping criteria, we combine the two
most obvious: i)~the maximum number of iterations; and ii)~the
minimum value of the norm of the update
vector~$\widehat{\bdelta}$. Using only ii) is not adequate in the
low resolution levels,  where it is only necessary to make a
coarse estimation of the parameters. In these levels, convergence
may be slow and the overall performance of the algorithm is not
affected if we simply perform a fixed number of iterations.

\subsection{Impact of the window}

The motion of the brightness pattern between two images $\bI$ and
$\bI'$ is described by the parametric mapping
$\bx'=\bm(\btheta,\bx)$ that maps each pixel of~$\bI$, with
coordinates $\bx$, into the corresponding pixel~$\bx'$ of~$\bI'$.
%,
%%. Under common assumptions relative to the geometry of
%%the scene and/or the camera motion, the global motion $\bm$ is
%%described by a small set of parameters~$\btheta$,
%{\it i.e.}, $\bx'=\bm(\bx,\btheta)$.
%
Featureless approaches to image registration estimate the global
motion parameter vector~$\btheta$ by minimizing the error
\begin{equation}
E(\btheta)\!=\!\sum_{\bx \in \cR} e^2(\btheta,\bx) ,\;\;\;\;
e(\btheta,\bx)\!=\!\bI(\bx)\!-\!\bI'(\bm(\btheta,\bx)) ,
\label{eq:funcional}
\end{equation}
where the sum is over a fixed, pre-specified, rectangular
window~$\cR$. When the overlap between the images is large, the
window~$\cR$ is simply chosen as a large rectangle in the interior
of the image(s). However, when the overlap is small, it is
difficult to select {\it a priori} an appropriate window $\cR$,
due to two reasons. First, since it is not known beforehand where
the overlapping region is, its hard to choose a location for the
window~$\cR$. Second, imposing {\it a priori} a small window,
leads to less accurate estimates of~$\btheta$ because not only the
minimum of~$E(\btheta)$ in~(\ref{eq:funcional}) becomes less sharp
but also the local minima phenomena become more severe.

\begin{figure}[hbt]
\centerline{\epsfig{figure=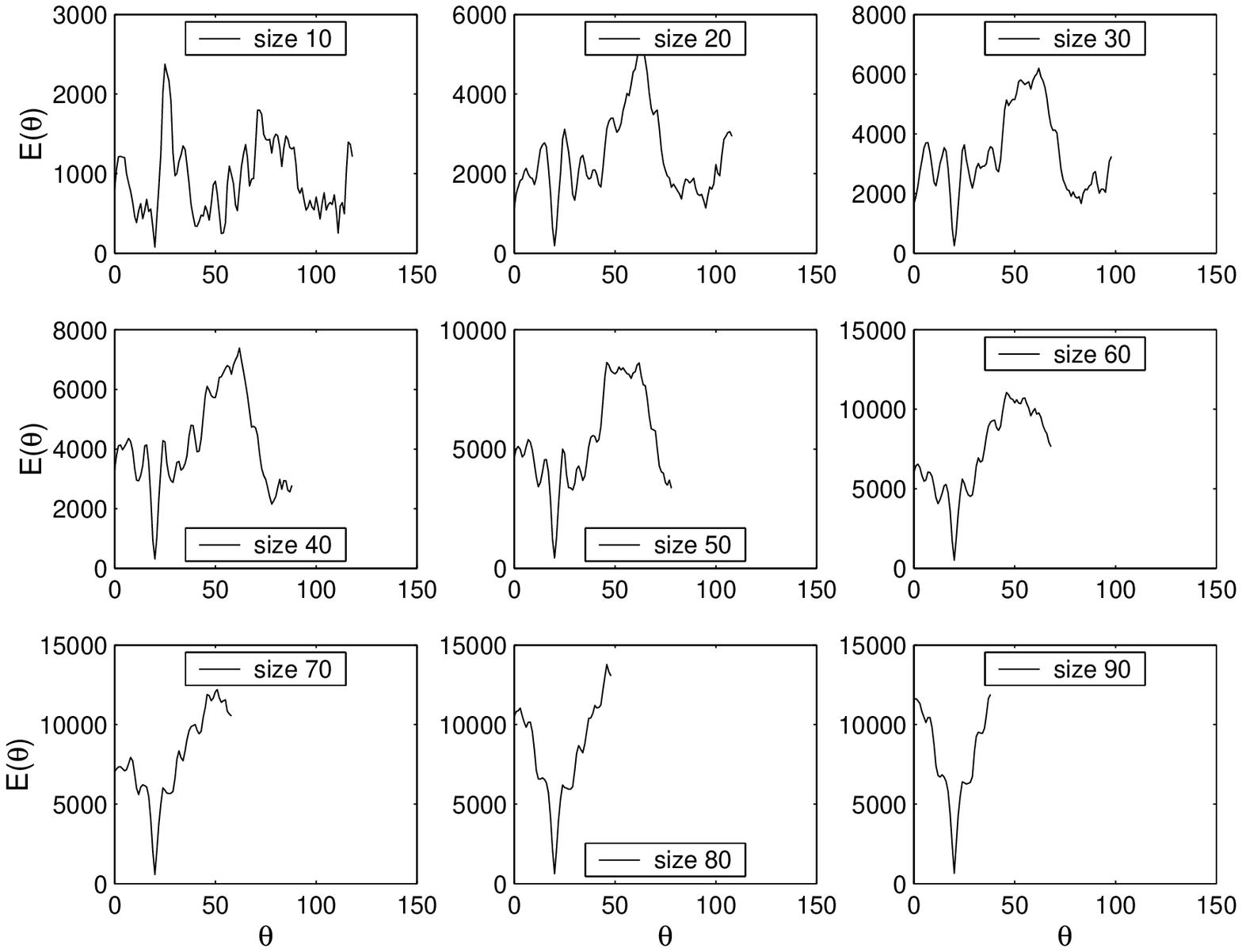,width=16cm}} \caption{Error
$E(\btheta)$ in~(\ref{eq:funcional}) for different sizes of the
window~$\cR$.\label{fig:window}}
\end{figure}

To illustrate the impact of the size of the window, we represent
in Fig.~\ref{fig:window} the typical evolution of $E$
in~(\ref{eq:funcional}), as a function of a single motion
parameter~$\theta$, for several sizes of~$\cR$. Naturally, as
anticipated above, the larger is $\cR$, the smaller is the
domain~$\{\theta\}$ in which $E(\theta)$ can be evaluated. The
several local minima and the smoothness of the minimum of
$E(\theta)$ at the true value $\theta\!=\!20$ in the top plots,
obtained with relatively small windows, contrast with the single
sharp minimum of the bottom-right plot, obtained with the largest
window (note that the vertical scale is different from plot to
plot).

\section{Multi-frame Registration}
\label{sec:mfr}

\subsection{Estimate of the registration parameters $\left\{\btheta_1,\ldots,\btheta_n\right\}$}

Replacing the ML estimate $\widehat{\bP}$ of the panorama, given
by~(\ref{eq:2}), in the symmetric log-likelihood~(\ref{eq:1}), we
express this ML~cost~$L$ as a function of the unknown registration
parameters $\bTheta$ alone. After algebraic manipulations, we get:
%\begin{equation}\label{eq:3}
%L\left(\bTheta\right)= \frac{nN}{2}\ln\left(2\pi\sigma^{2}\right)
%+ \left(4\sigma^{2}\right)^{-1}\!\!\sum_{\bx_0\in\R^2}
%\bW^{-1}\!\left(\bx_0,\bTheta\right) \sum_{i,j=1}^n
%\bE_{ij}^2\left(\bx_0,\btheta_i,\btheta_j\right)\,
%\bH\left(\bm(\btheta_i;\bx_0)\right)\,
%\bH\left(\bm(\btheta_j;\bx_0)\right)\,,\!\!\!\!\!\!\!\!\!\!\!\!\!\!\!\!\nonumber
%\end{equation}
\begin{eqnarray}\label{eq:3}
L\left(\bTheta\right)= \frac{nN}{2}\ln\left(2\pi\sigma^{2}\right)
+ \left(4\sigma^{2}\right)^{-1}\!\!\sum_{\bx_0\in\R^2}
\bW^{-1}\!\left(\bx_0,\bTheta\right) \cdot&&\\&& \hspace*{-7.25cm}
\cdot\sum_{i,j=1}^n
\bE_{ij}^2\left(\bx_0,\btheta_i,\btheta_j\right)\,
\bH\left(\bm(\btheta_i;\bx_0)\right)\,
\bH\left(\bm(\btheta_j;\bx_0)\right)\,,\!\!\!\!\!\!\!\!\!\!\!\!\!\!\!\!\nonumber
\end{eqnarray}
where $\bE_{ij}$ is the error between the co-registered images $\bI_i$
and $\bI_j$,
\begin{equation}
\bE_{ij}\left(\bx_0,\btheta_i,\btheta_j\right)=\label{eq:e}
\bI_i\left(\bm(\btheta_i;\bx_0)\right)-\bI_j\left(\bm(\btheta_j;\bx_0)\right)\,,
\end{equation}
and $\bW\left(\bx_0,\bTheta\right)$ is a weight that counts the
number of images that have captured the pixel~$\bx_0$ of the
panorama, according to the registration parameters in $\bTheta$,
{\it i.e.},
\begin{equation} \bW\left(\bx_0,\bTheta\right)=\sum_{k=1}^n
\bH\left(\bm(\btheta_k;\bx_0)\right)\,.\label{eq:w}
\end{equation}

By discarding from~(\ref{eq:3}) the constant terms, {\it i.e.},
the terms that do not depend on the unknown registration
parameters~$\bTheta$, we conclude that the ML estimate for the
problem of global multi-frame registration, is equivalent to the
following minimization:
\begin{equation}
\widehat{\bTheta}=\arg\min_{\bTheta} \sum_{i,j=1}^n
\sum_{\bx_0\in\cR_{ij}}
\frac{\bE_{ij}^2\left(\bx_0,\btheta_i,\btheta_j\right)}{\bW\left(\bx_0,\bTheta\right)}\,.
\label{eq:4}\end{equation} For simplicity, when
deriving~(\ref{eq:4}) from~(\ref{eq:3}),  the sums were
interchanged and the spatial region of summation was re-defined to
take care of the windows~$\bH(\cdot)$ in~(\ref{eq:3}), {\it i.e.},
$\cR_{ij}$ in~(\ref{eq:4}) is the region where the images $\bI_i$
and $\bI_j$ overlap,
\begin{equation}
\cR_{ij}=\left\{\bx:\bH\left(\bm(\btheta_i;\bx)\right)\,\bH\left(\bm(\btheta_j;\bx)\right)=1\right\}\,.
\end{equation}

Expressions~(\ref{eq:w}) and~(\ref{eq:4}) condense one the
contributions of this paper---they show that the ML estimate
$\widehat{\bTheta}$ of the registration parameters $\bTheta$ is
given by the minimum of a particular weighted sum of the square
differences between all possible pairs of co-registered input
images.

\section{MLM Algorithm}
\label{sec:mlma}

Our algorithm to the minimization of the ML~cost~(\ref{eq:4}) uses
an iterative scheme inspired in the common approaches to the
two-frame problem~\cite{mann97,pires04}. In each step, the
algorithm updates a current estimate that we denote
by~$\bTheta^0=\left\{\btheta_1^0,\ldots,\btheta_n^0\right\}$.

\subsection{Iterative minimization of the ML cost}

Instead of updating the entire set of parameters $\bTheta$ in a
single step, which would be computationally complex, we propose a
coordinatewise minimization: we update each vector~$\btheta_q$ at
a time, keeping fixed the remaining registration parameters
$\left\{\btheta_i\!=\btheta_i^0,i\neq q \right\}$. The update is
$\btheta_q\!=\btheta_q^0+\widehat{\bdelta}$, where
$\widehat{\bdelta}$ is obtained from~(\ref{eq:4}), after
discarding the terms that do not depend on $\btheta_q$:
\begin{equation}
\widehat{\bdelta}=\arg\min_{\bdelta} \sum_{i=1}^n
\sum_{\bx_0\in\cR_{iq}}\!\!
\frac{\bE_{iq}^2(\bx_0,\btheta_i^0,\btheta_q^0+\bdelta)}
{\bW\left(\bx_0,\bTheta^0\right)} \label{eq:5}
\end{equation}

To obtain a closed-form solution for the update
$\widehat{\bdelta}$, we approximate the error $\bE_{iq}$ by its
first-order Taylor series expansion,
\begin{equation}
\bE_{iq}(\bx_0,\btheta_i^0,\btheta_q^0+\bdelta)\approx
\bE_{iq}(\bx_0,\btheta_i^0,\btheta_q^0)+
\bdelta^T\cdot\nabla_{\!\btheta_{\!
q}}\!\!\bE_{iq}(\bx_0,\btheta_i^0,\btheta_q^0)\,.\nonumber
\end{equation}
From the definition of $\bE_{iq}$ in~(\ref{eq:e}), the gradient in
the Taylor series expansion is easily computed in terms of the
spatial gradient of image~$\bI_q$. Furthermore, that gradient does
not depend on $\theta_i^0$, thus we will denote it more compactly
by $\nabla(\bx_0,\btheta_q^0)$,
\begin{eqnarray}\label{eq:gradsimple}
\nabla(\bx_0,\btheta_q^0) &\!\!\!\!=\!\!\!\!&
\nabla_{\!\btheta_q}\!\bE_{iq}(\bx_0,\btheta_i^0,\btheta_q^0)\\
&\!\!\!\!=\!\!\!\!&
-\nabla_{\!\btheta_q}\!\bm(\btheta_q^0;\bx_0)\cdot
\nabla_{\bx}\bI_q(\bm(\btheta_q^0;\bx_0))\,.\;
\end{eqnarray}

By inserting the Taylor series approximation in~(\ref{eq:5}) and
making zero the derivative wrt $\bdelta$, we get the update
$\widehat{\bdelta}$ as the solution of a linear system
\begin{equation}\label{eq:ls}
\bGamma\left(\bTheta^0\right)\cdot\widehat{\bdelta}+\bgamma\left(\bTheta^0\right)=\bzero\,.
\end{equation}
The matrix~$\bGamma\!\left(\bTheta^0\right)$ and the
vector~$\bgamma\left(\bTheta^0\right)$ are obtained~as
%\!\!\!\!\!\!\!\!\!\!\!\!\!\!\!\!\!\!\!\!\!\!\!\!
\begin{eqnarray}\label{eq:gamma1}
\!\!\!\!\!\!\!\!\bGamma\left(\bTheta^0\right)\!\!\!\!\!&=&\!\!\!\!\!\!\!\sum_{\bx_0\in\cR_q}
\nabla(\bx_0,\btheta_q^0) \cdot \nabla^T(\bx_0,\btheta_q^0)\,,\\
%\end{equation}
%\begin{equation}
\!\!\!\!\!\!\!\!\bgamma\left(\bTheta^0\right)\!\!\!\!\!&=&\!\!\!\!\!\!\!\sum_{\bx_0\in\cR_q}\!\!
\nabla(\bx_0,\btheta_q^0)\!\left[\widehat{\bP}\left(\bx_0,\!\bTheta^0\right)
\!-\! \bI_q(\bm(\btheta_q^0;\bx_0))\right],\label{eq:gamma2}
\end{eqnarray}
where we used expression~(\ref{eq:2}) for $\widehat{\bP}$. The
sums in~(\ref{eq:gamma1},\ref{eq:gamma2}) are over the region
observed by image~$\bI_q$,
$\;\cR_q\!=\!\left\{\bx:\bH\left(\bm(\btheta_q^0;\bx)\right)\!=\!1\right\}$.

%\begin{eqnarray}
%\bdelta^T \cdot \left(\sum_{\bx\in\bR^2} \nabla_{\btheta^w}
%e_w(\bx^0,\btheta_0^w) \cdot \nabla_{\btheta^w}^T
%e_w(\bx^0,\btheta_0^w) \cdot \bh^w(\bx^0,\btheta^w) \right) +\nonumber\\
%\sum_{\bx\in\bR^2} \nabla_{\btheta^w}^T\, e_w(\bx^0,\btheta_0^w)
%\left[\widehat{P(\bx^0)}- I^w(m(\bx^0,\btheta\,^w)) \right] \cdot
%\bh^w(\bx^0,\btheta^w) =\bzero
%\end{eqnarray}

\subsection{Interpretation in terms of current algorithms}

Since the iterations in standard featureless two-frame alignment
algorithms~\cite{mann97,pires04} also lead to a system
like~(\ref{eq:ls}), we now interpret our
solution~(\ref{eq:ls},\ref{eq:gamma1},\ref{eq:gamma2}) in terms of
those approaches. Define $\bE_{0q}$ as the difference between
image~$\bI_q$ and the previous estimate of the panorama, obtained
with the registration parameters~$\bTheta^0$,
\begin{equation}
\bE_{0q}\left(\bx_0,\bTheta^0\right)=\widehat{\bP}\left(\bx_0,\bTheta^0\right)-
\bI_q(\bm(\btheta_q^0;\bx_0))\,. \end{equation} Since the gradient
of this error wrt $\btheta_q$ is equal to the one defined
in~(\ref{eq:gradsimple}),
%\begin{equation}
%$\nabla_{\btheta_q}\!\bE_{0q}\left(\bx_0,\bTheta^0\right)=\nabla(\bx_0,\btheta_q^0)$,
%\,.
%\end{equation}
we can re-write expressions (\ref{eq:gamma1},\ref{eq:gamma2})
 in terms of $\bE_{0q}$,
%\begin{eqnarray}
%\nabla_{\btheta^w}\be_{0w}(\bx^0,\bTheta_0) = \nabla_{\btheta^w}^T
%e_w(\bx^0,\btheta_0^w) =\\ -\nabla_{\btheta^w}
%\bm(\bx^0,\btheta_0^w)\cdot\nabla_{\bx^w} I'(\bx^w)
%\end{eqnarray}
\begin{eqnarray}\label{eq:gamma3}
\!\!\!\!\!\!\!\!\bGamma\left(\bTheta^0\right)\!\!\!\!&=&\!\!\!\!\!\!\sum_{\bx_0\in\cR_q}
\nabla_{\btheta_q}\!\bE_{0q}\left(\bx_0,\bTheta^0\right) \cdot
\nabla_{\btheta_q}^T\!\bE_{0q}\left(\bx_0,\bTheta^0\right)\,,\\
\!\!\!\!\!\!\!\!\bgamma\left(\bTheta^0\right)\!\!\!\!&=&\!\!\!\!\!\!\sum_{\bx_0\in\cR_q}
\nabla_{\btheta_q}\!\bE_{0q}\left(\bx_0,\bTheta^0\right)\,\,\bE_{0q}\left(\bx_0,\bTheta^0\right)\,.
\label{eq:gamma4}
\end{eqnarray}

%\begin{eqnarray}
%\bdelta^T \cdot \left(\sum_{\bx\in\bR^2} \nabla_{\btheta^w}
%e_0w(\bx^0,\bTheta_0) \cdot \nabla_{\btheta^w}^T
%e_0w(\bx^0,\bTheta_0) \cdot \bh^w(\bx^0,\btheta^w)
%\right)\nonumber\\ + \sum_{\bx\in\bR^2} \nabla_{\btheta^w}^T\,
%e_0w(\bx^0,\bTheta_0) e_0w(\bx^0,\bTheta_0) \cdot
%\bh^w(\bx^0,\btheta^w) =\bzero
%\end{eqnarray}

Expressions~(\ref{eq:gamma3},\ref{eq:gamma4}) are equal to the
ones that arise from aligning the previous estimate
$\widehat{\bP}$ of the panorama with image~$\bI_q$, by using
standard featureless methods, see {\it e.g.},
\cite{mann97,pires04} or \cite{aguiar01-icip}. We thus conclude
that our global approach lead to an algorithm that refines the
estimate of the registration parameters of each image by using the
methodology developed to register a single pair of images.
%Note that
%this equivalence comes from the fact that the panorama is
%estimated using ML (\ref{eq:2}).

\subsection{Convergence---initialization and multiresolution}

Our algorithm starts by aligning the images sequentially, using
the standard two-frame approach~\cite{mann97,pires04}. Then, we
compute an initial estimate of the panorama by using~(\ref{eq:2}).
After this, we cyclically refine the registrations parameters of
each image. The stopping criterion may either be the error below a
small threshold or reaching a maximum number of iterations.

%This means that the parameters of any picture can be refined even
%if the picture as already been aligned with the panoramic image
%which enables us to combat error propagation when aligning sets of
%images.

Since the truncated Taylor series is a good approximation only
when the vector $\btheta_q$ is close to its initial value
$\btheta_q^0$, estimating the update~$\bdelta$ from
(\ref{eq:ls},\ref{eq:gamma1},\ref{eq:gamma2}) leads to the
convergence to the globally optimal ML estimate, only when the
initial estimate is close enough to it. However, in practice, {\it
e.g.}, in the first experiment described below, it is common that
the initial estimate of the panorama is very rough, due to the
propagation of (two-frame based) registration errors. To cope with
these situations, we use a coarse-to-fine  approach similar to the
one proposed in~\cite{bergen92,mann97}: the parameters are first
estimated in the coarsest resolution level, then used as an
initialization to the next finer level, until the full image
resolution is attained. As illustrated in the following section,
this multi-resolution approach succeeds in correcting large
miss-registrations.

\section{Experiments}
\label{sec:exp}

We describe two experiments. The first experiment compares our
global approach with the current sequential registration methods.
In the second experiment, we illustrate with automatic mosaic
building in a seabed mapping context.

\subsection{Adaptive window versus fixed window}

To illustrate how our method performs better than the usual fixed
window method, we synthesized input images by cropping a real
photography and adding noise. This corresponds to a simple
translational motion model, which suffices to show the advantage
of using our adaptive window method.
%
%. This is a simplification of the affine model where the matrix
%$\bA$ is made the identity matrix.
%
%
%To make it easier to obtain experimental results and compare both
%fixed and adaptive methods we use only one real image. Portions of
%this image were isolated and noise was added synthetically. This
%was to guarantee that the images could be precisely registered
%with the translation model.

\begin{figure}[hbt]
\centerline{\epsfig{figure=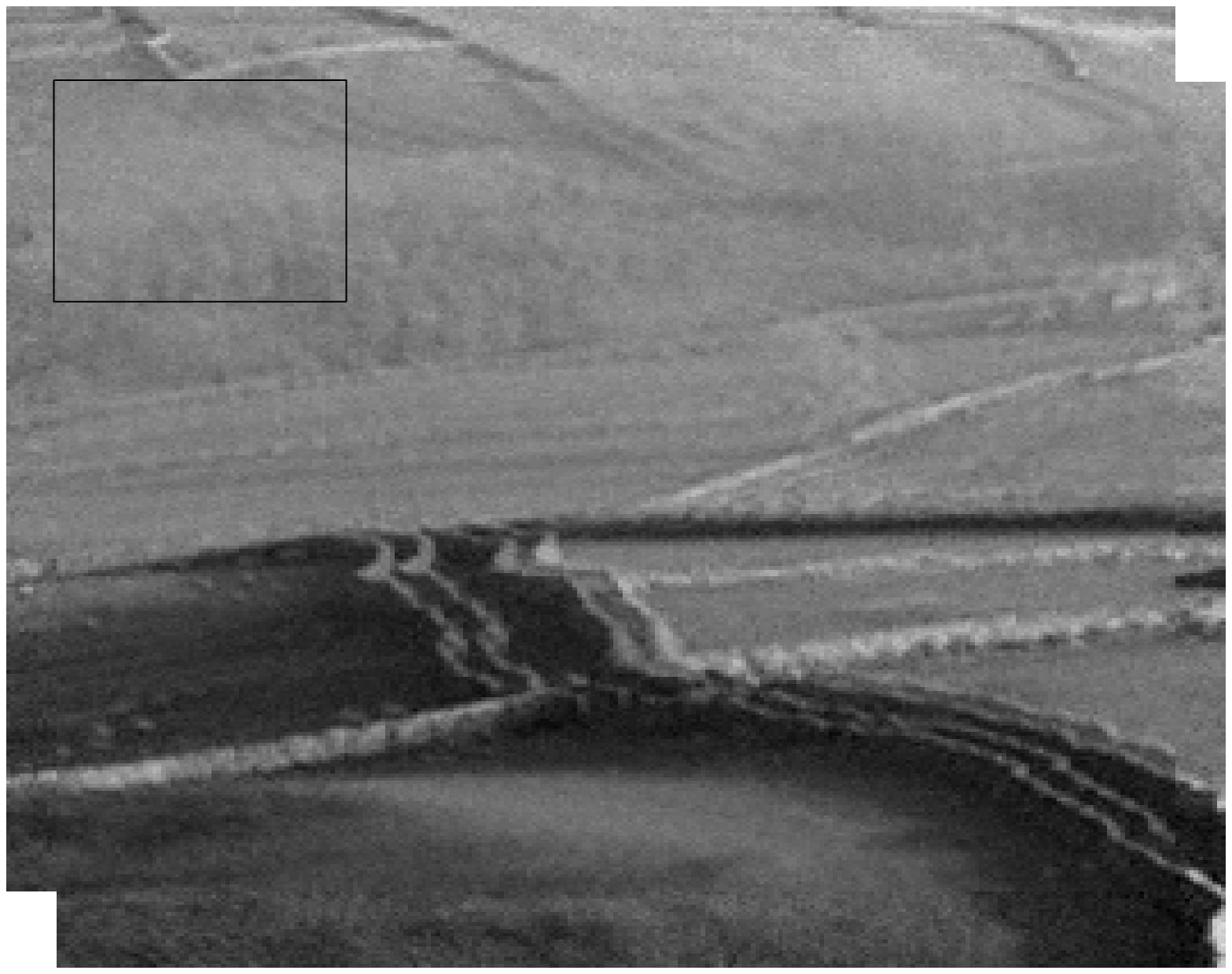,width=5cm} \ \
\epsfig{figure=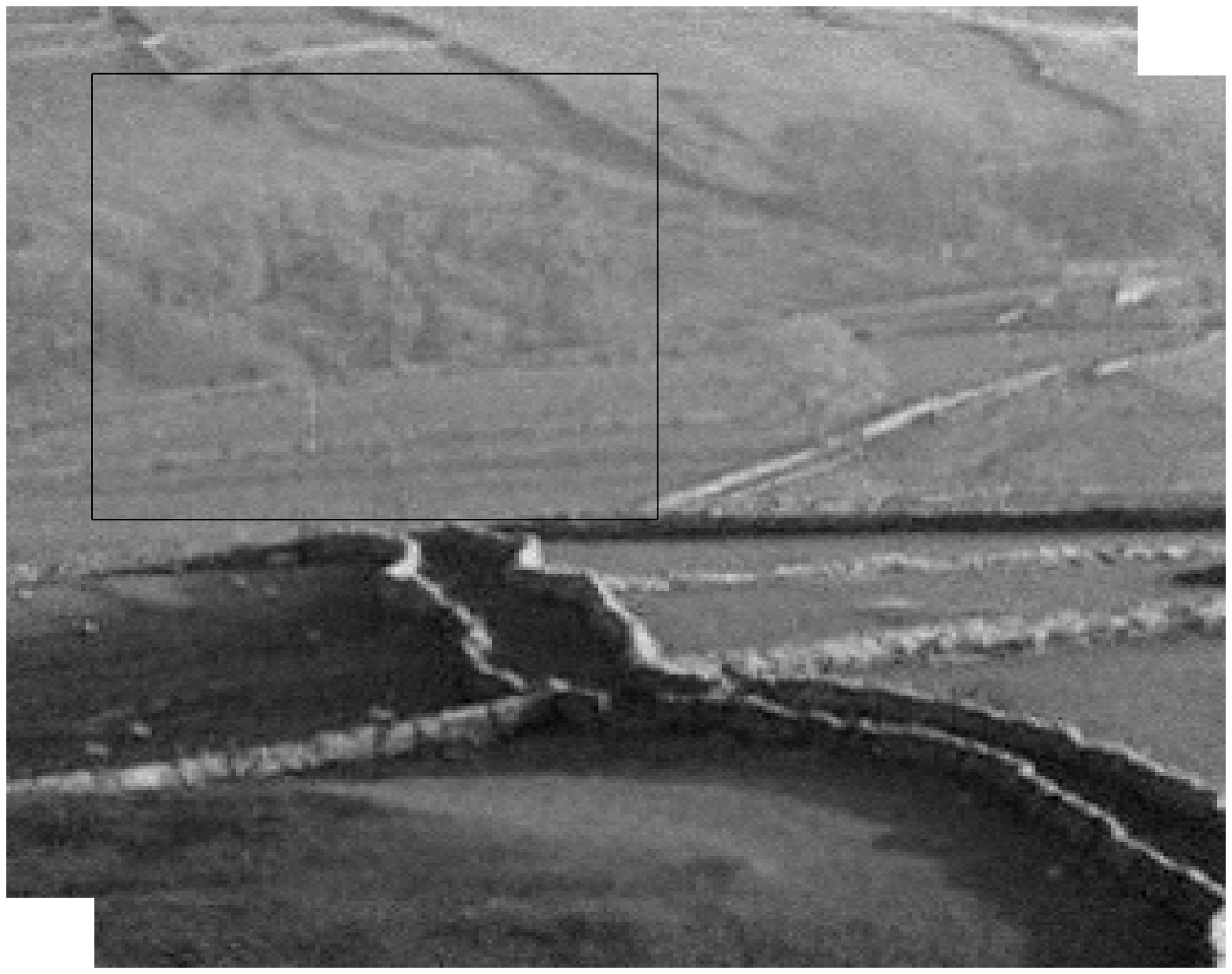,width=5cm}\ \
\epsfig{figure=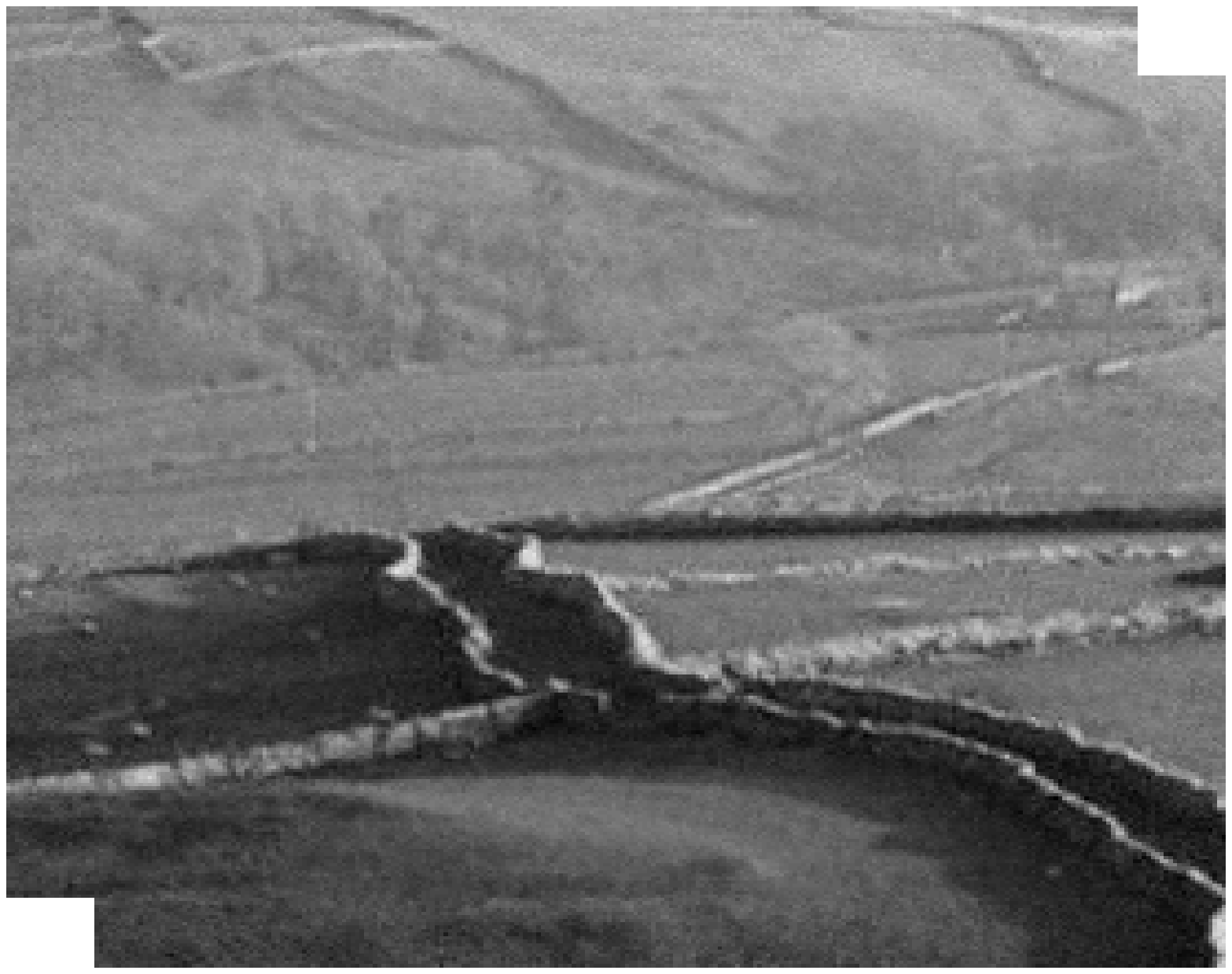,width=5cm}} \caption{Registration of a
pair of images. Left: using a fixed window of size~64
(registration failure). Middle: fixed window size~128. Right: our
algorithm.\label{fig:1}}
\end{figure}

In Fig.~\ref{fig:1}, the overlap between input images is large.
The left image of Fig.~\ref{fig:1} shows the failure of the
algorithm with a fixed window of size 64. Our algorithm and the
one with a fixed window of size 128 both lead to good results, see
the middle and right images of Fig.~\ref{fig:1}. Note that,
although these two images are visually indistinguishable, the
estimate of the global motion provided by our algorithm is more
accurate because it minimizes the error over the largest possible
window.

\begin{figure}[hbt]
\centerline{\epsfig{figure=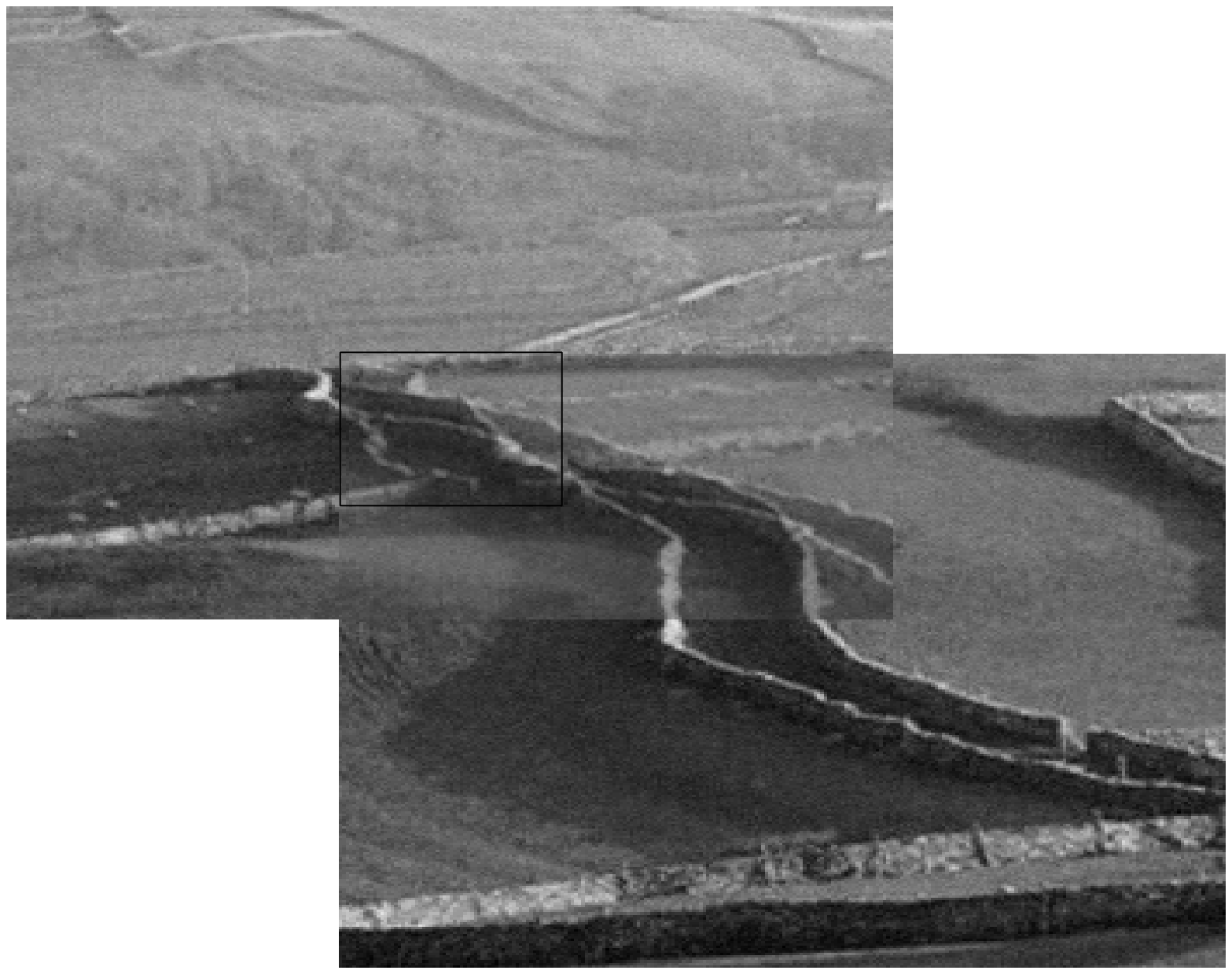,width=7.5cm,height=6.5cm}
\epsfig{figure=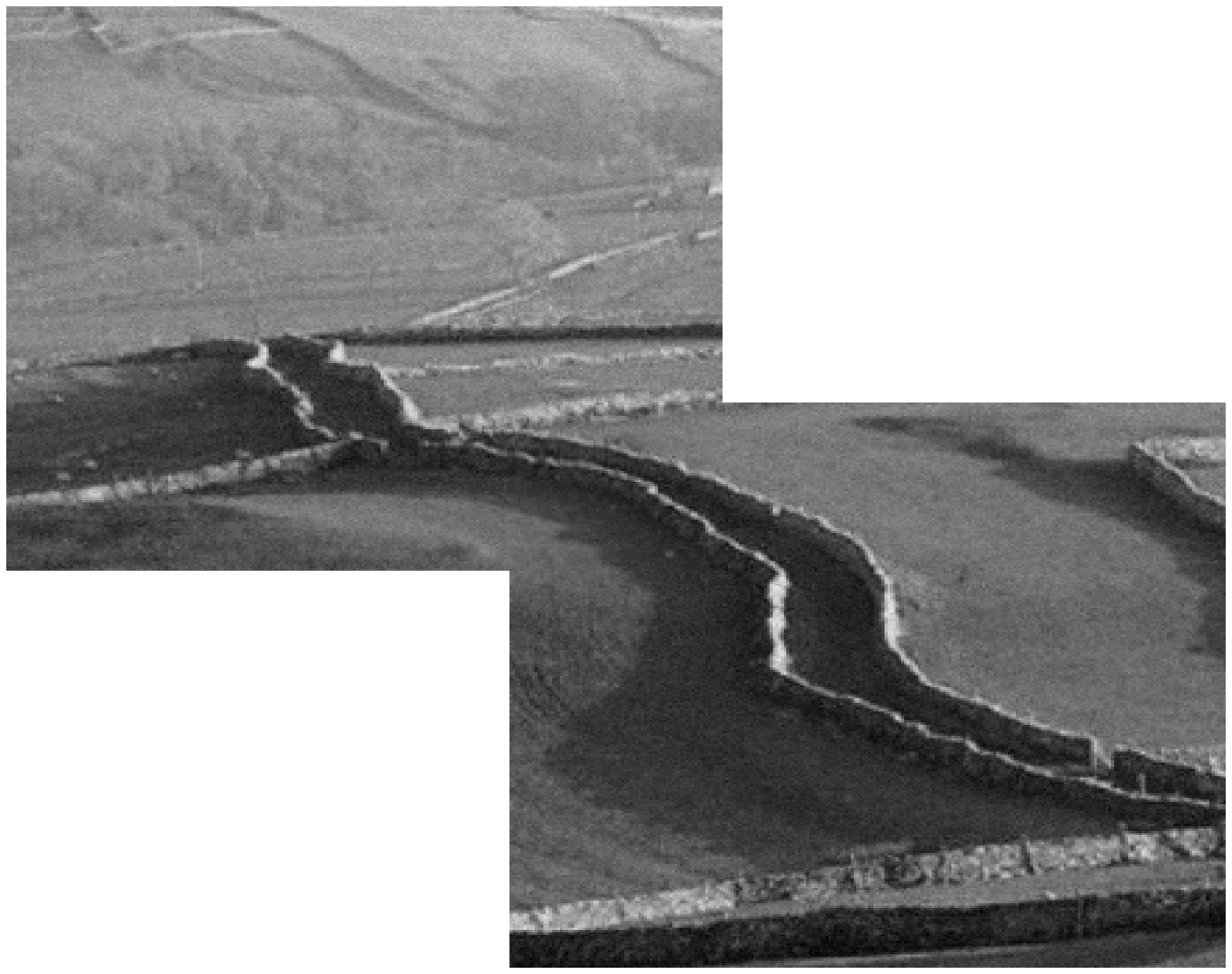,width=9cm}} \caption{Registration of
images with very small overlap. Left: using a fixed window of
size~64 (registration failure). Right: our
algorithm.\label{fig:2}}
%\caption{Registration of images with very
%small overlap. Left: fixed widow, size 64 (registration failure).
%Right: our algorithm.\label{fig:2}}
\end{figure}

In Fig.~\ref{fig:2}, the overlap between input images is small,
thus it is impossible to use a fixed window of a large size. When
using a fixed window of size 64, the usual algorithm fails, see
the left image of Fig.~\ref{fig:2}. The right image of
Fig.~\ref{fig:2} shows that our algorithm succeeds in this
challenging situation.

\begin{figure}[hbt]
\centerline{\epsfig{figure=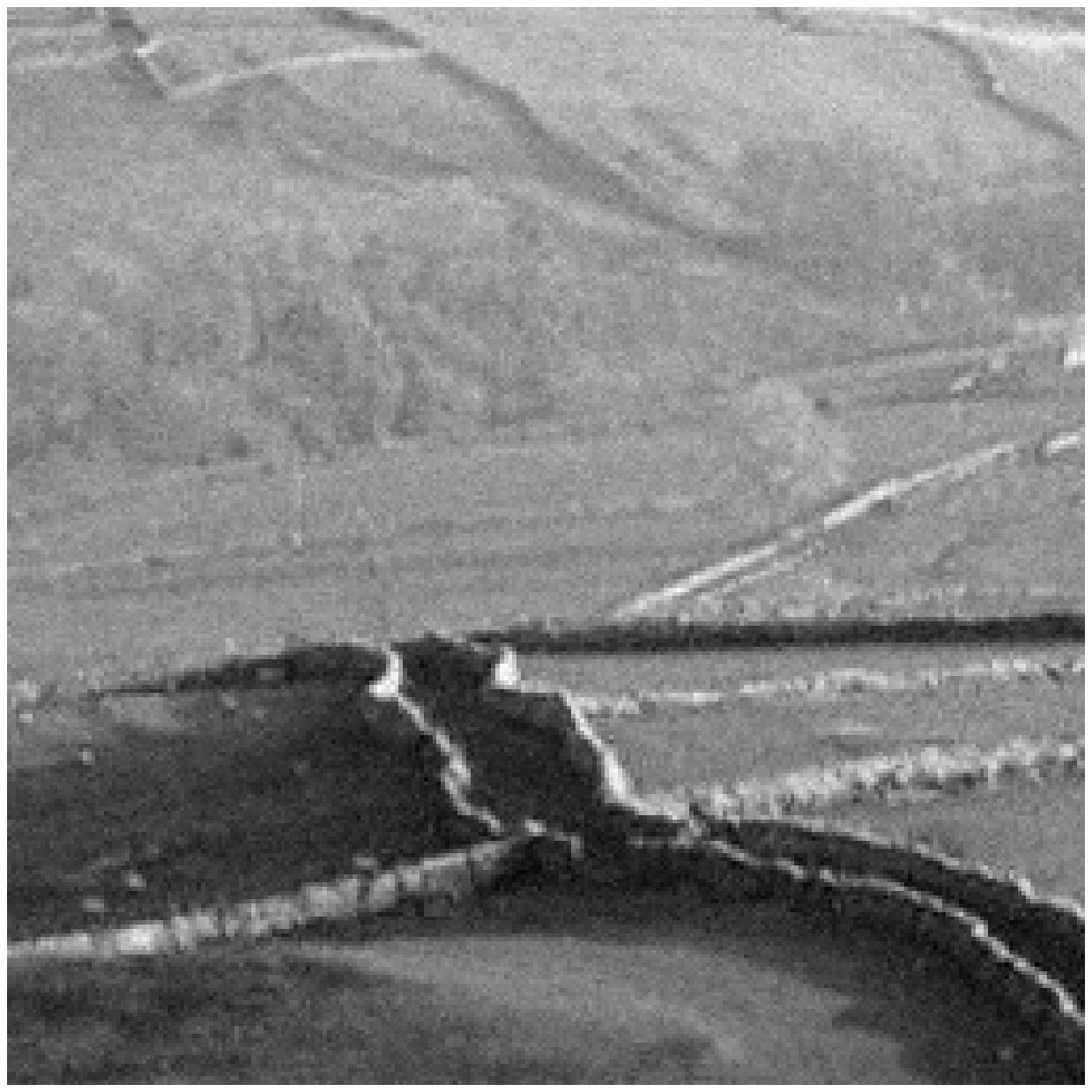,width=4cm,height=3cm}
\epsfig{figure=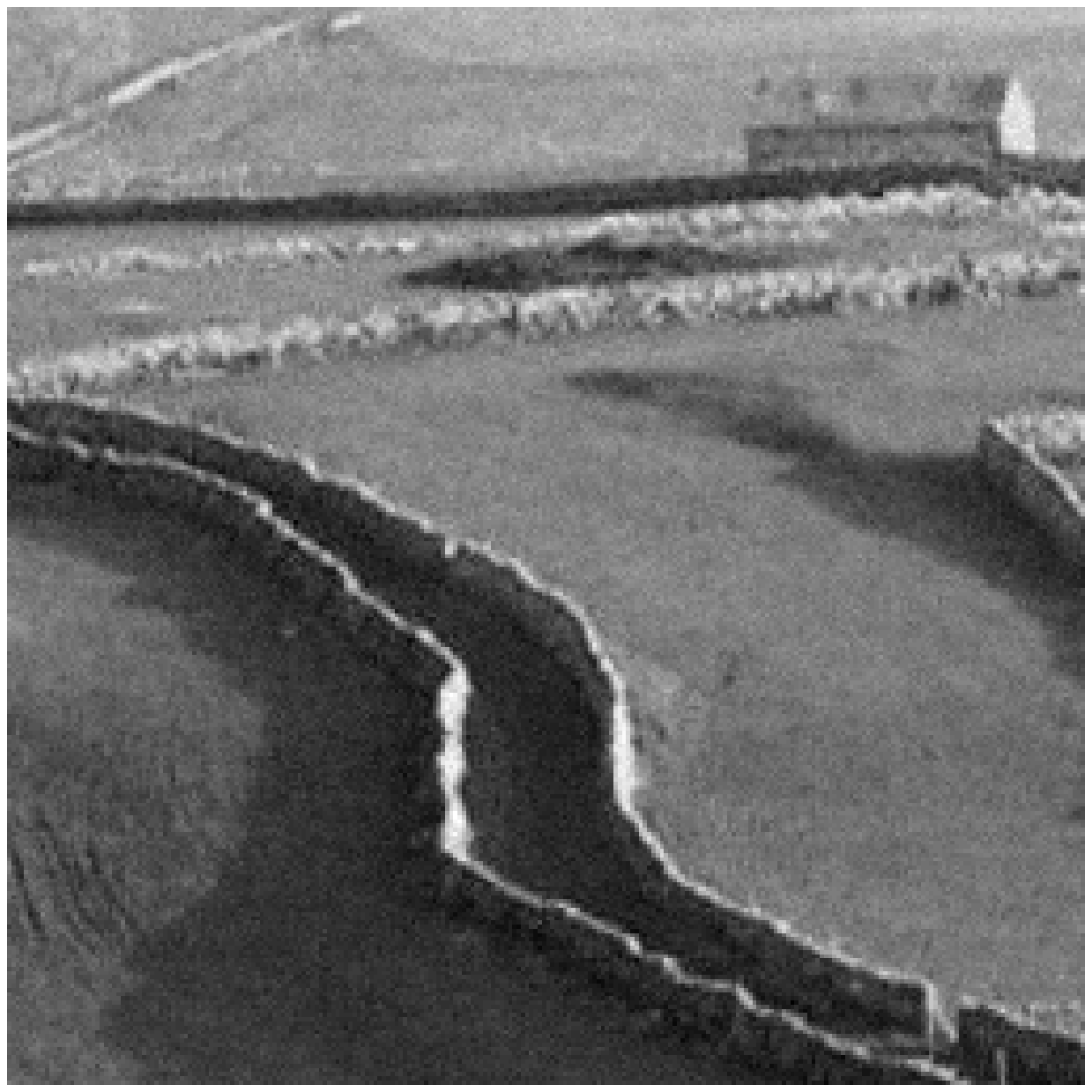,width=4cm,height=3cm}
\epsfig{figure=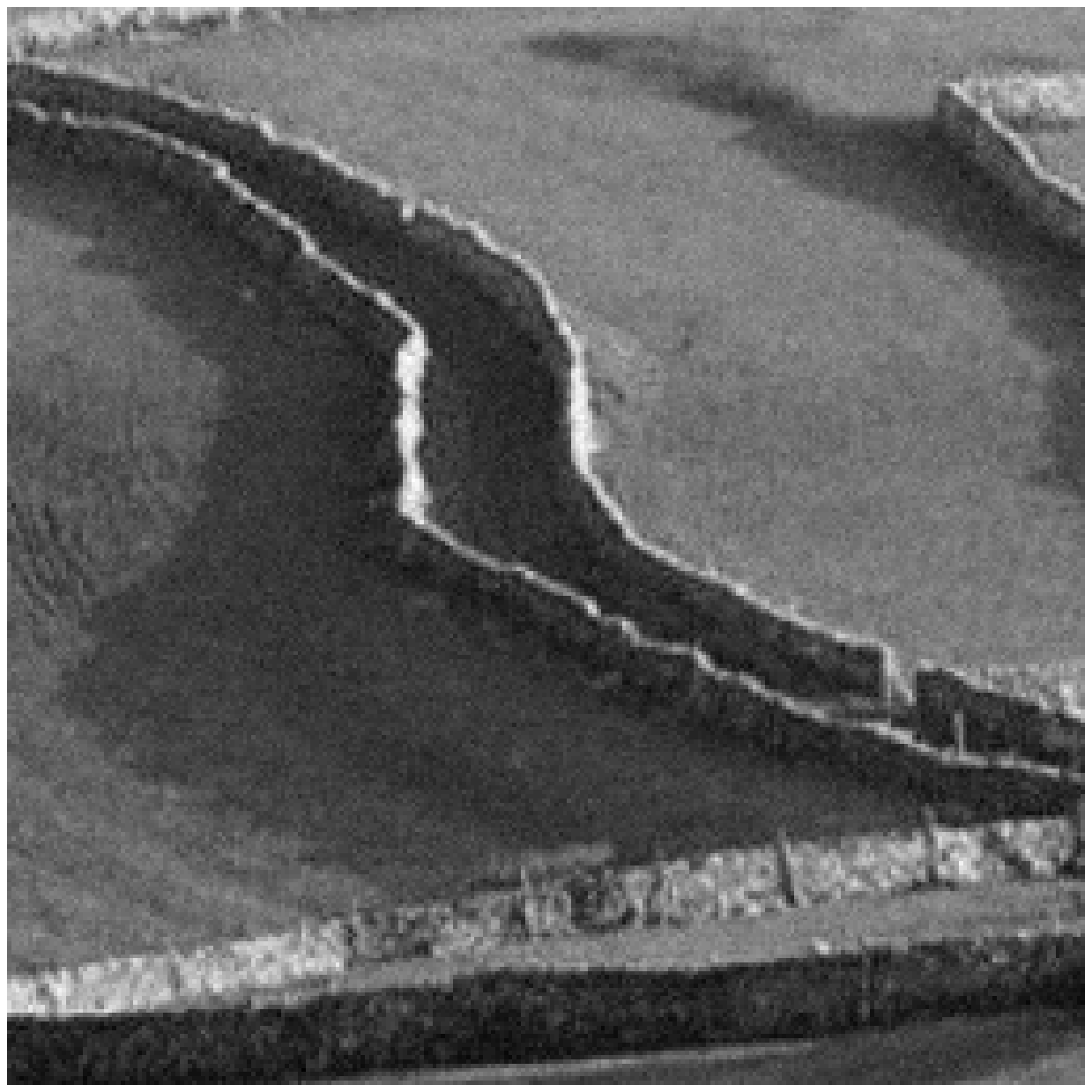,width=4cm,height=3cm}
\epsfig{figure=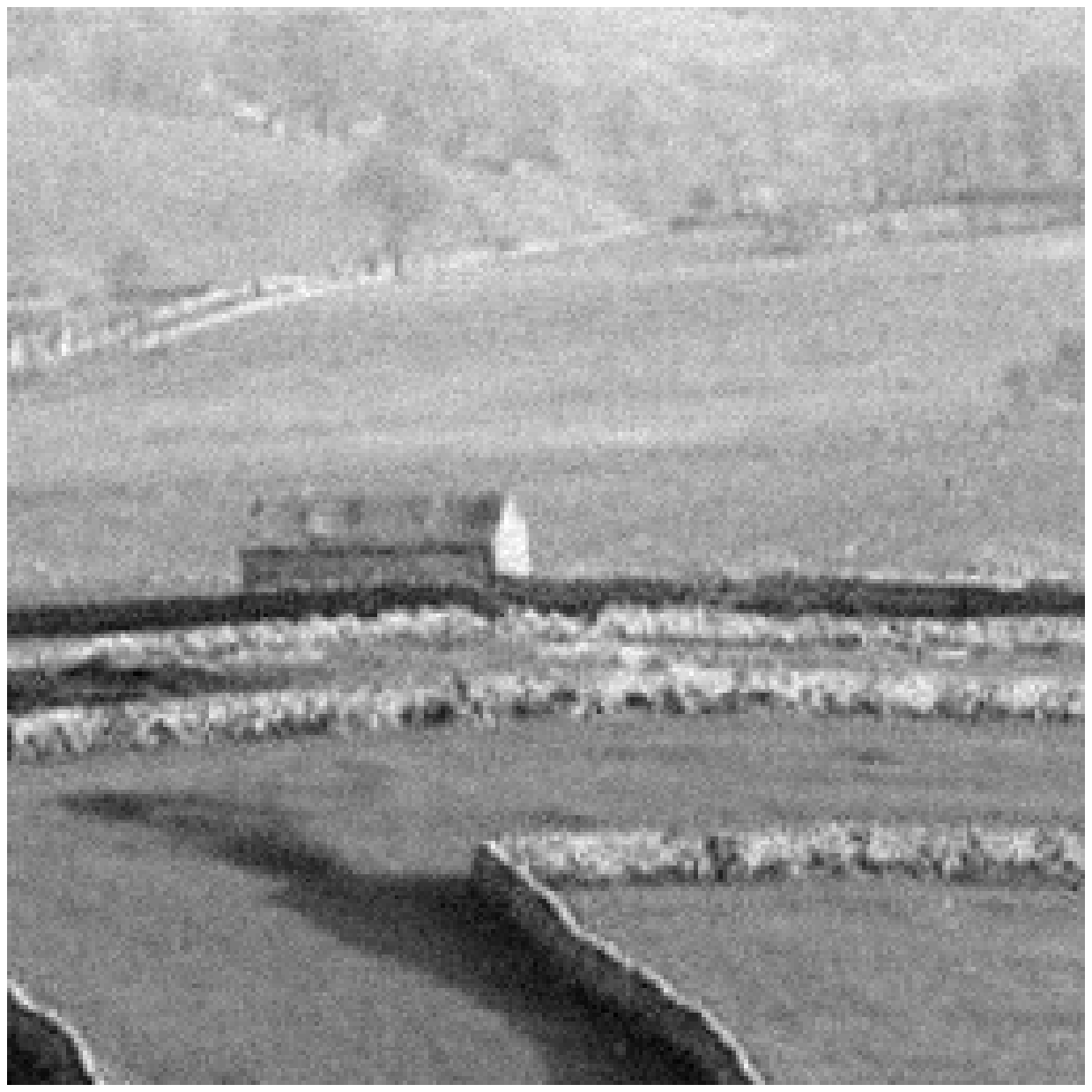,width=4cm,height=3cm} }\vspace*{.25cm}
%\centerline{\epsfig{figure=panorama_final.eps,width=7cm,height=4cm}}
\caption{Mosaic of images with small overlap. Top: original
images. Bottom: mosaic built by using our algorithm to register
those images.} \label{fig:3}
\end{figure}

Finally, Fig.~\ref{fig:3} shows a mosaic obtained by pairwise
registering, sequentially, a set of input images.

%Finally Fig.\ref{fig:3} shows the proper registration of the five
%different images shown in Fig.\ref{fig:4}. In order to achieve
%this result the images were registered in pairs sequentially and
%then a final panoramic image was created using all the information
%available.
%

\subsection{MLM versus sequential alignment}

\begin{figure}[htb]
\centerline{\psfig{figure=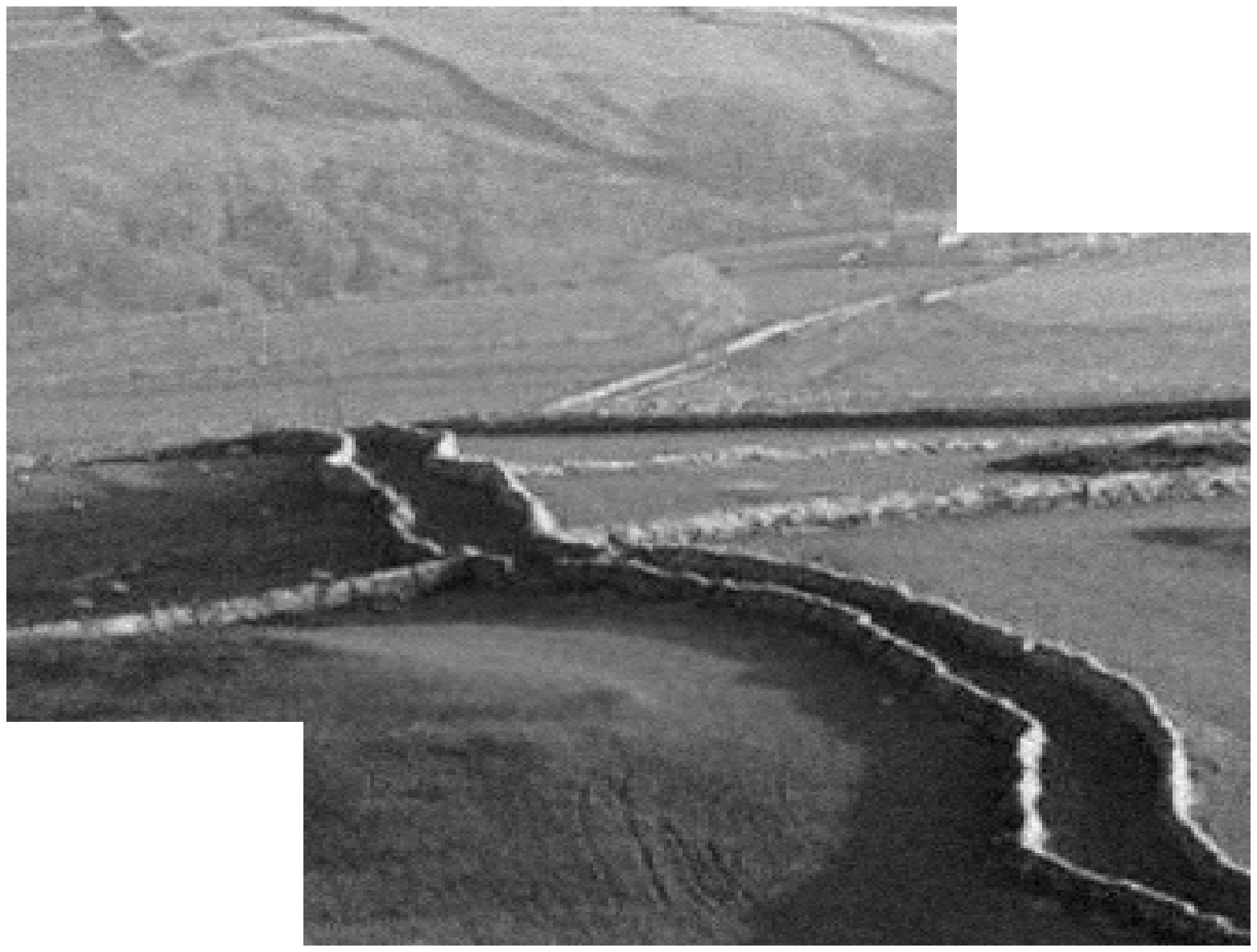,width=8cm} \hspace*{.15cm}
\psfig{figure=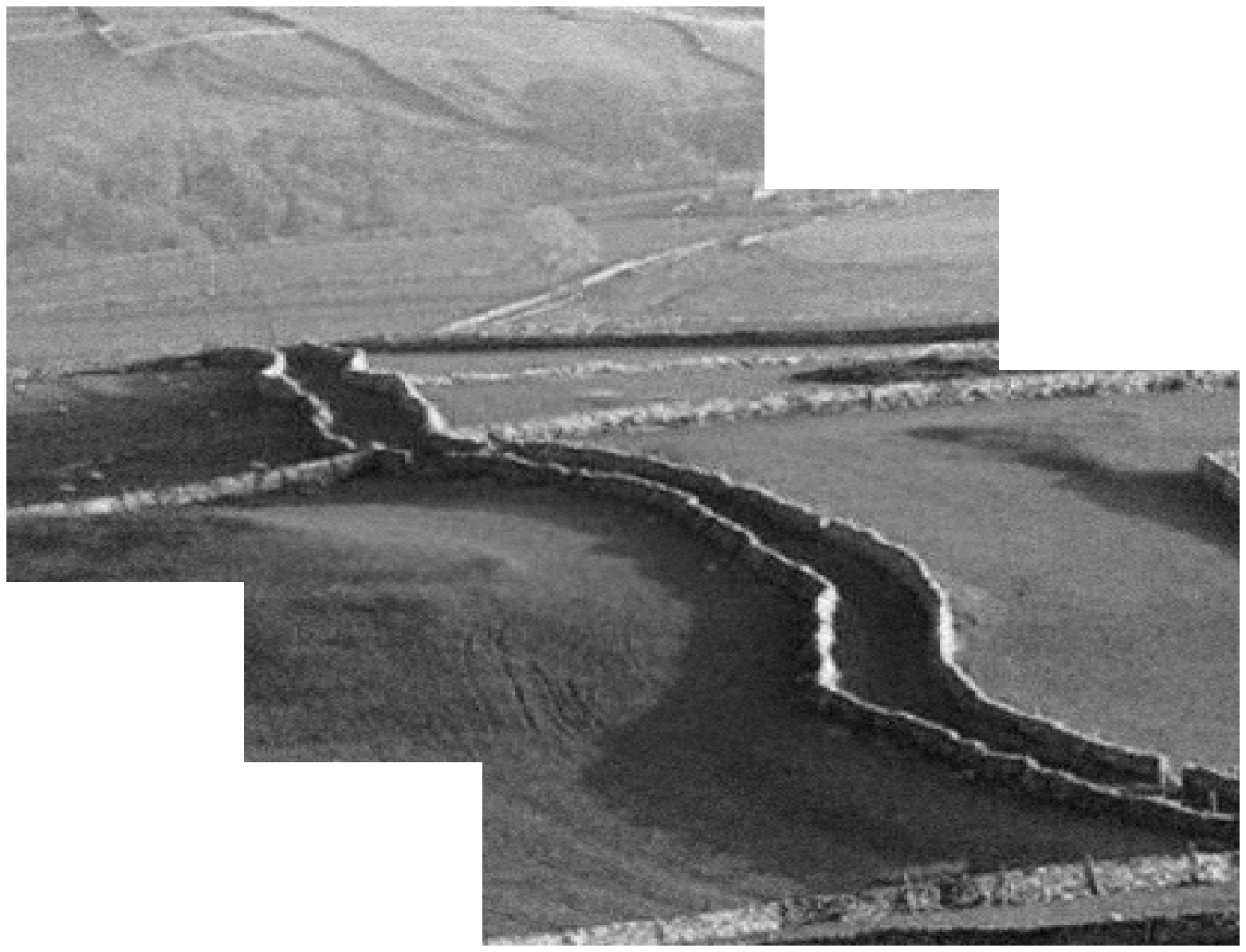,width=8cm}}
\centerline{\psfig{figure=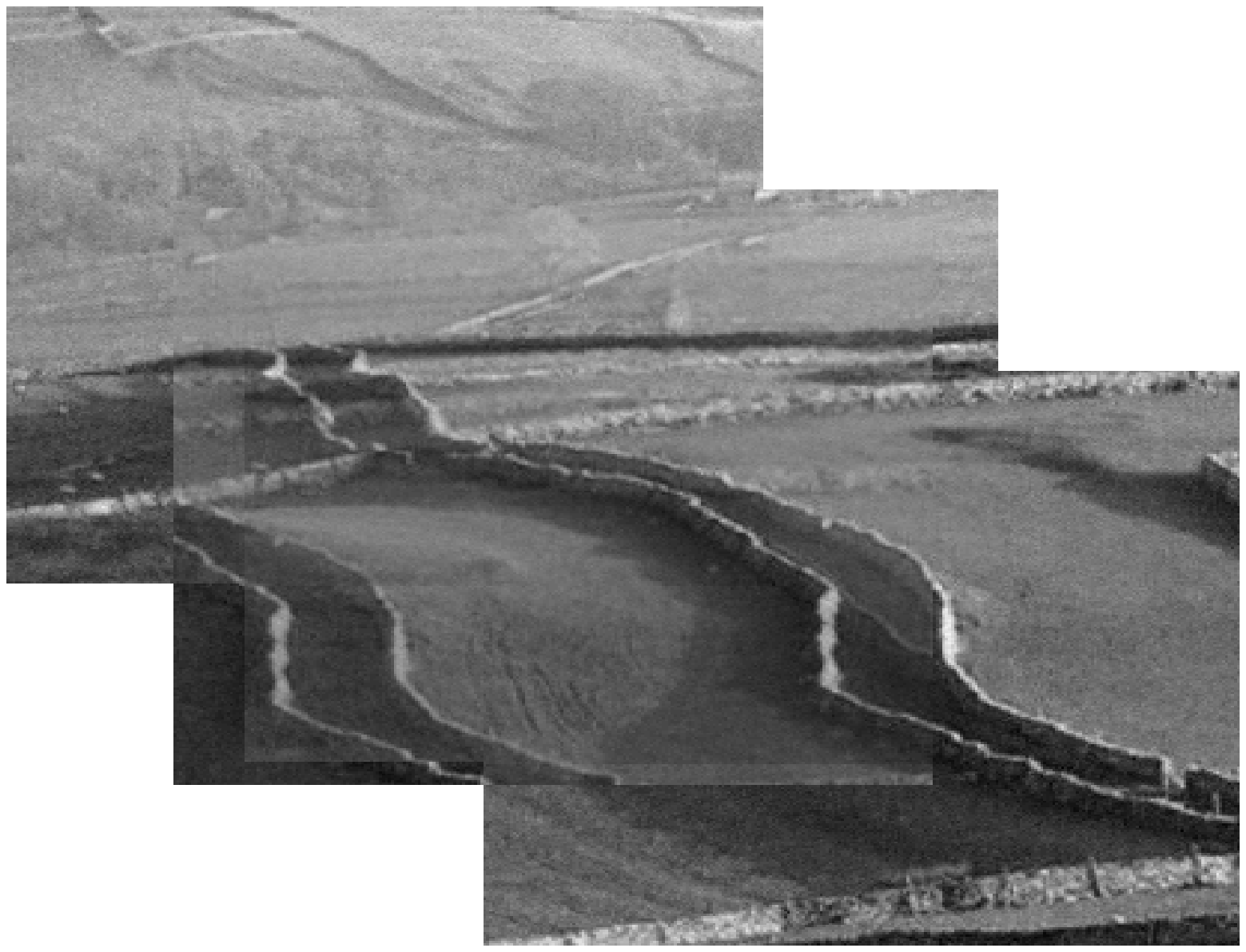,width=8cm} \hspace*{.15cm}
\psfig{figure=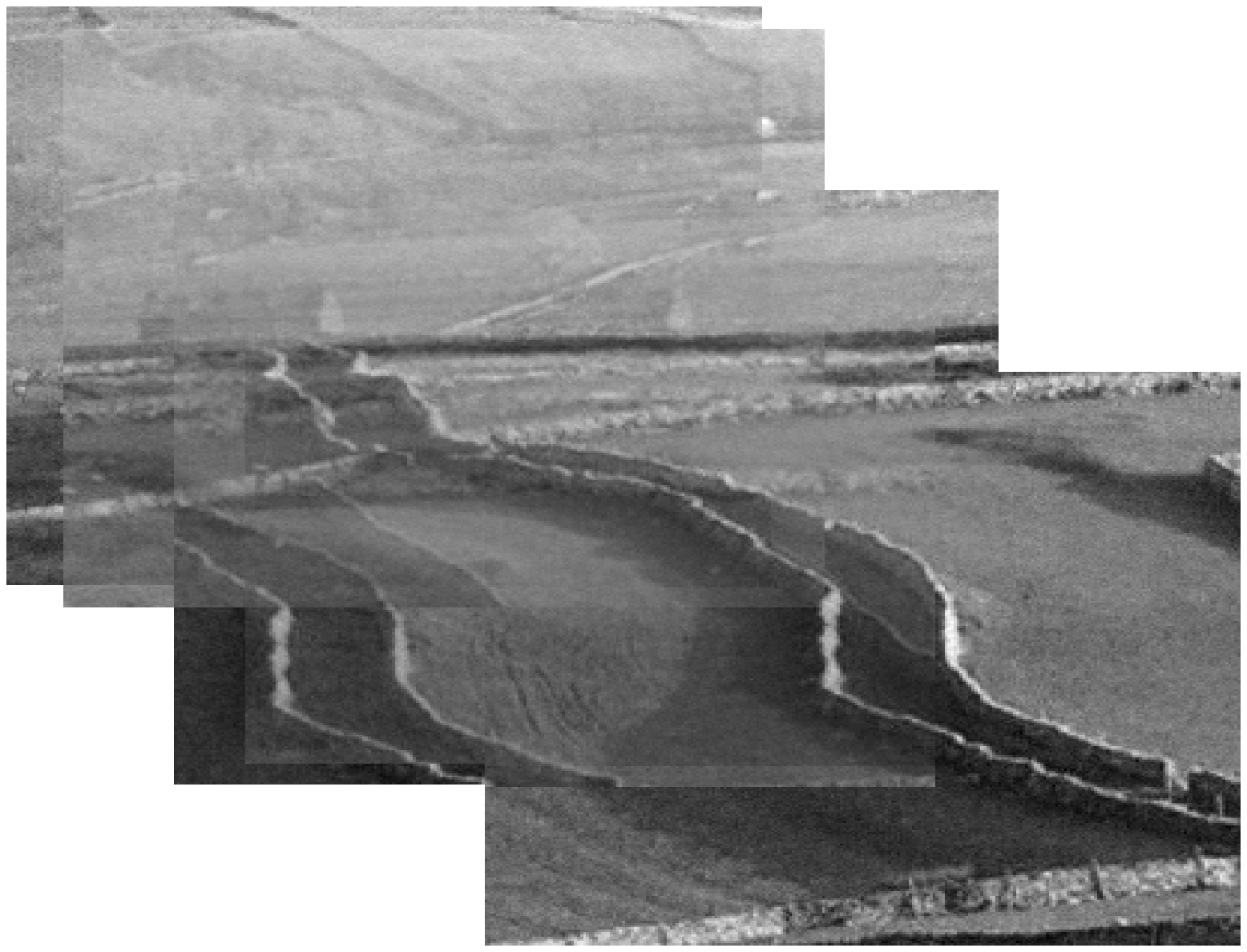,width=8cm}}
\centerline{\psfig{figure=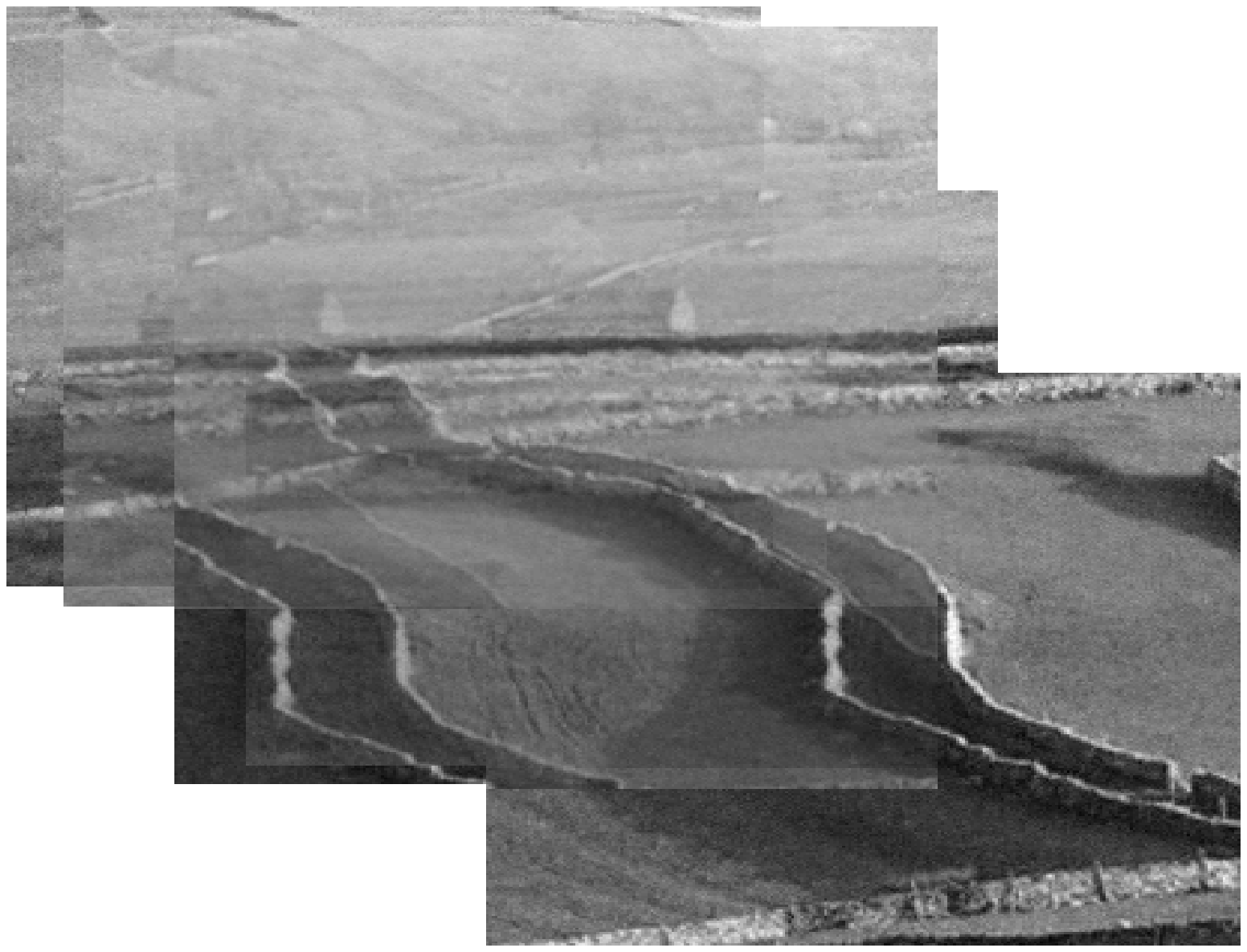,width=8cm} \hspace*{.15cm}
\psfig{figure=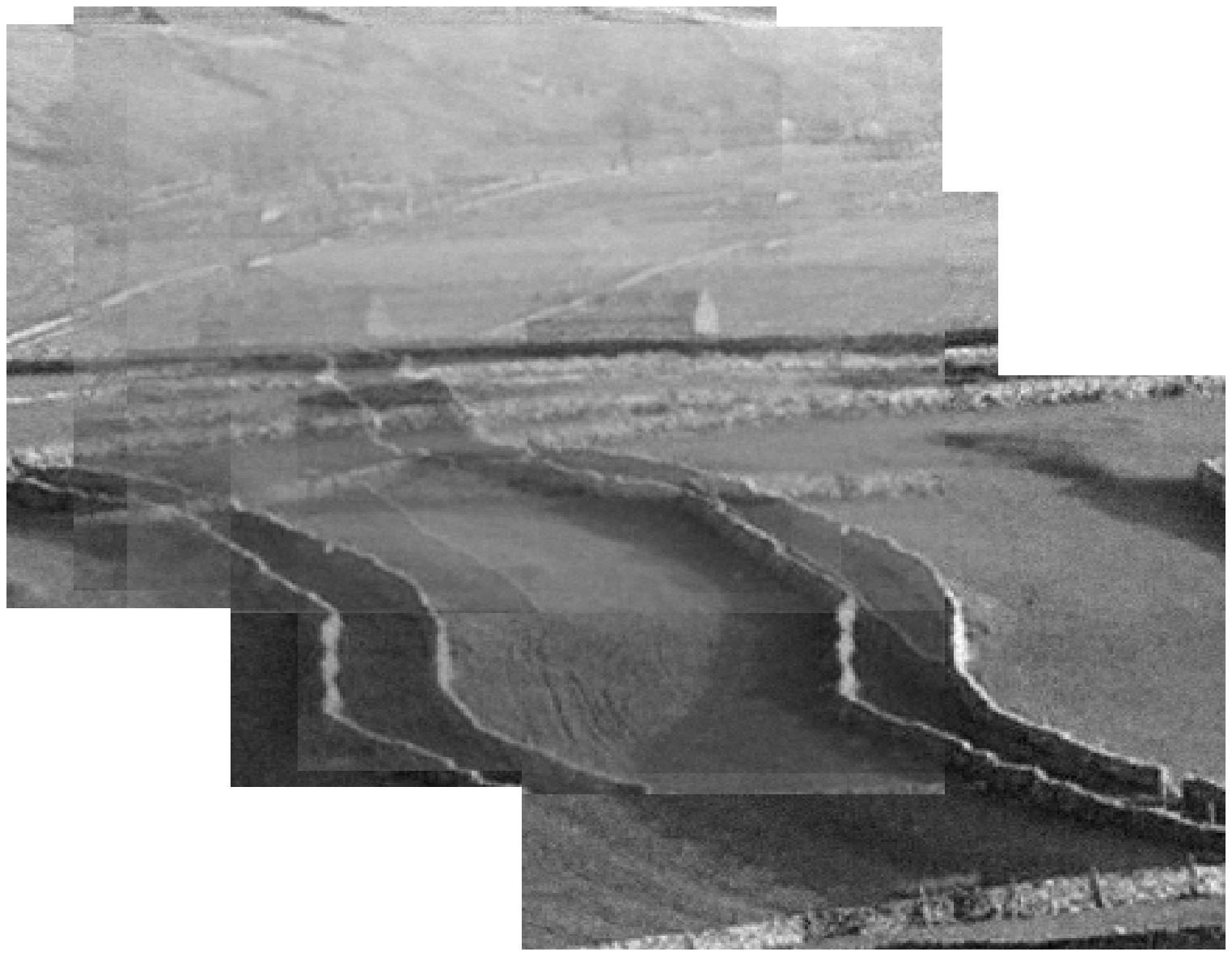,width=8cm}} \caption{Sequential
registration. Note how the miss-alignment of the fourth image
(middle left) propagates to the remaining ones.\label{fig:_2}}
\end{figure}

\begin{figure}[htb]
\centerline{\psfig{figure=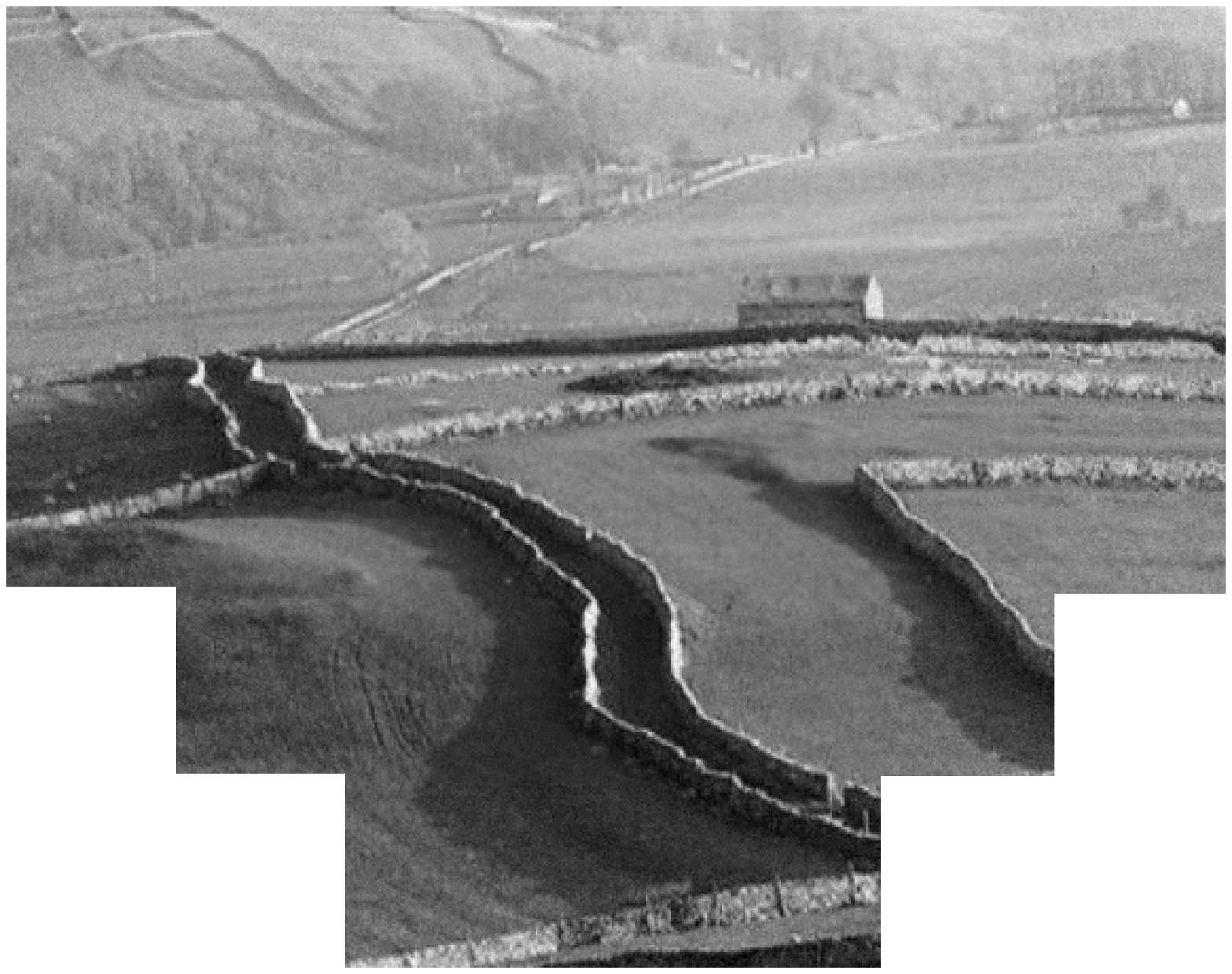,width=14cm}}
\caption{Proposed approach. Final estimate of the panorama, when
our algorithm is initialized with the bottom right image of
Fig.~\ref{fig:_2}.\label{fig:_3}}
\end{figure}

%This section shows, through a simple experiment, the superiority
%of the proposed method over the the simple sequential alignment.
To have an exact knowledge of the ground truth, we ``synthesized"
the input images by cropping a real photo and adding noise.
%, this
%guarantees that the images can be aligned using a simple
%translation model. Despite the simplicity of this experiment it
%suffices to show the superiority of the proposed method.
In Fig.~\ref{fig:_2}, we represent the evolution of the standard
two-frame featureless sequential alignment ({\it e.g.},
\cite{mann97,pires04}) of those images. Note that the fourth image
is miss-aligned and how that error propagates to the alignment of
the remaining images. The (highly incorrect) panorama this way
obtained, see the bottom right image of Fig.~\ref{fig:_2}, was
then used as the initialization for the global method we propose
in this paper. After few iterations, our algorithm converged to
the panoramic image shown in Fig.~\ref{fig:_3}, which is visually
indistinguishable from the ground truth image.

\subsection{Underwater mosaic for seabed mapping}\label{subsec:3}

As a final example, we illustrate our method with automatic mosaic
construction from video images, captured by an underwater camera
in the sea. Although underwater images are particularly difficult
to align, due to the absence of salient features, the mosaic
recovered by our algorithm is visually correct,
see~Fig.~\ref{fig:mosaic2}.

%\begin{figure}[hbt]
%\centerline{\epsfig{figure=under_1.eps,width=4cm}
%\epsfig{figure=under_2.eps,width=4cm}}\vspace*{.1cm}
%\centerline{\epsfig{figure=under_3.eps,width=4cm}
%\epsfig{figure=under_4.eps,width=4cm}} \caption{Sample underwater
% images.} \label{fig:mosaic}
%\end{figure}

%\begin{figure}[hbt]
%\centerline{$\;\;\begin{array}{cc} \begin{array}{cc}
%\psfig{figure=under_1.eps,width=1.8cm}&\!\!\!\!
%\psfig{figure=under_2.eps,width=1.8cm}\\
%\psfig{figure=under_3.eps,width=1.8cm}&\!\!\!\!
%\psfig{figure=under_4.eps,width=1.8cm}\end{array}
%&\!\!\!\!\!\!\!\!\!\!\begin{array}{cc}\\\psfig{figure=under_1234.eps,width=4.5cm}\end{array}\end{array}$}
%\caption{Seabed mapping: video frames and underwater mosaic.}
%\label{fig:mosaic2}
%\end{figure}

\begin{figure}[hbt]
\centerline{\epsfig{figure=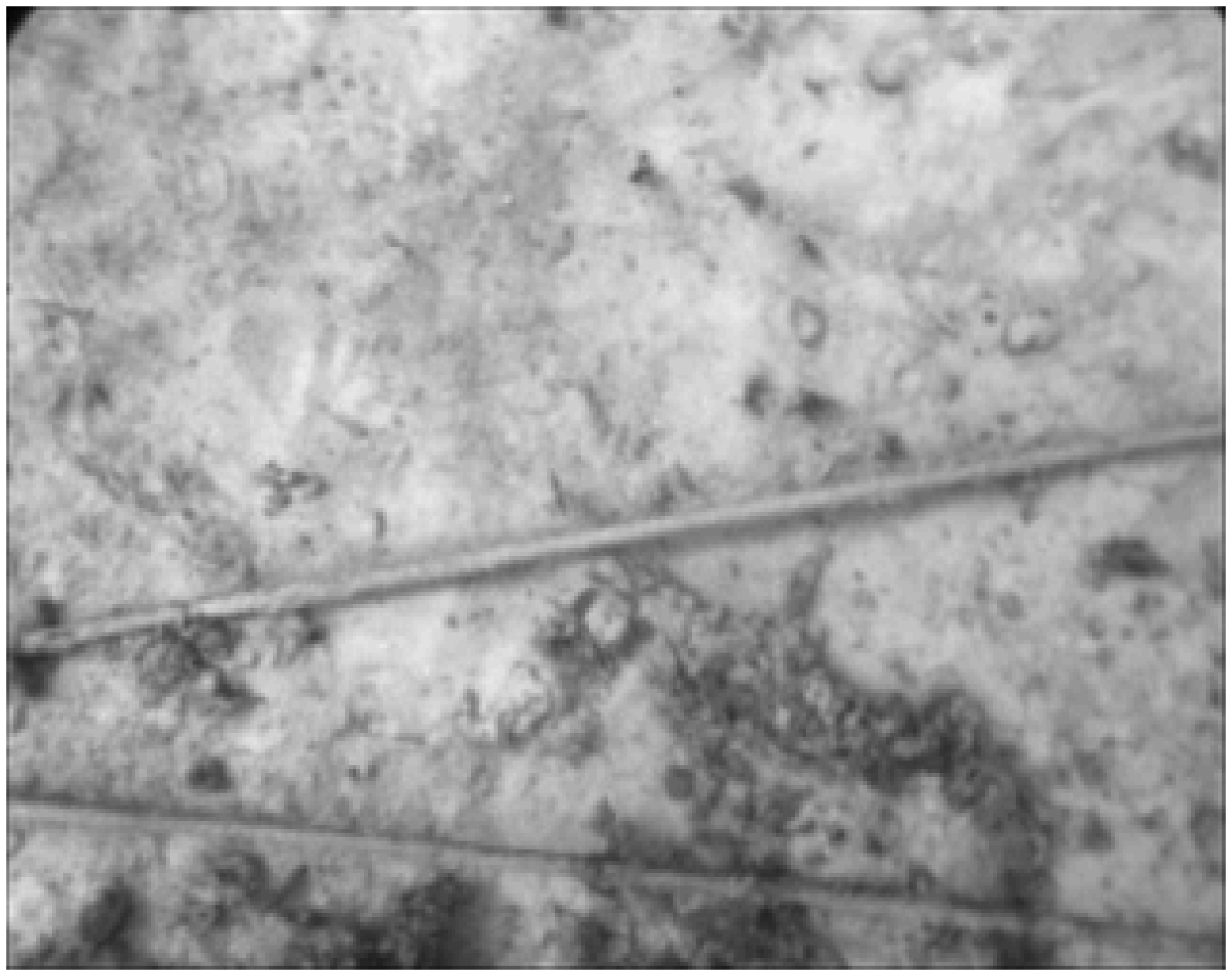,width=4cm}
\epsfig{figure=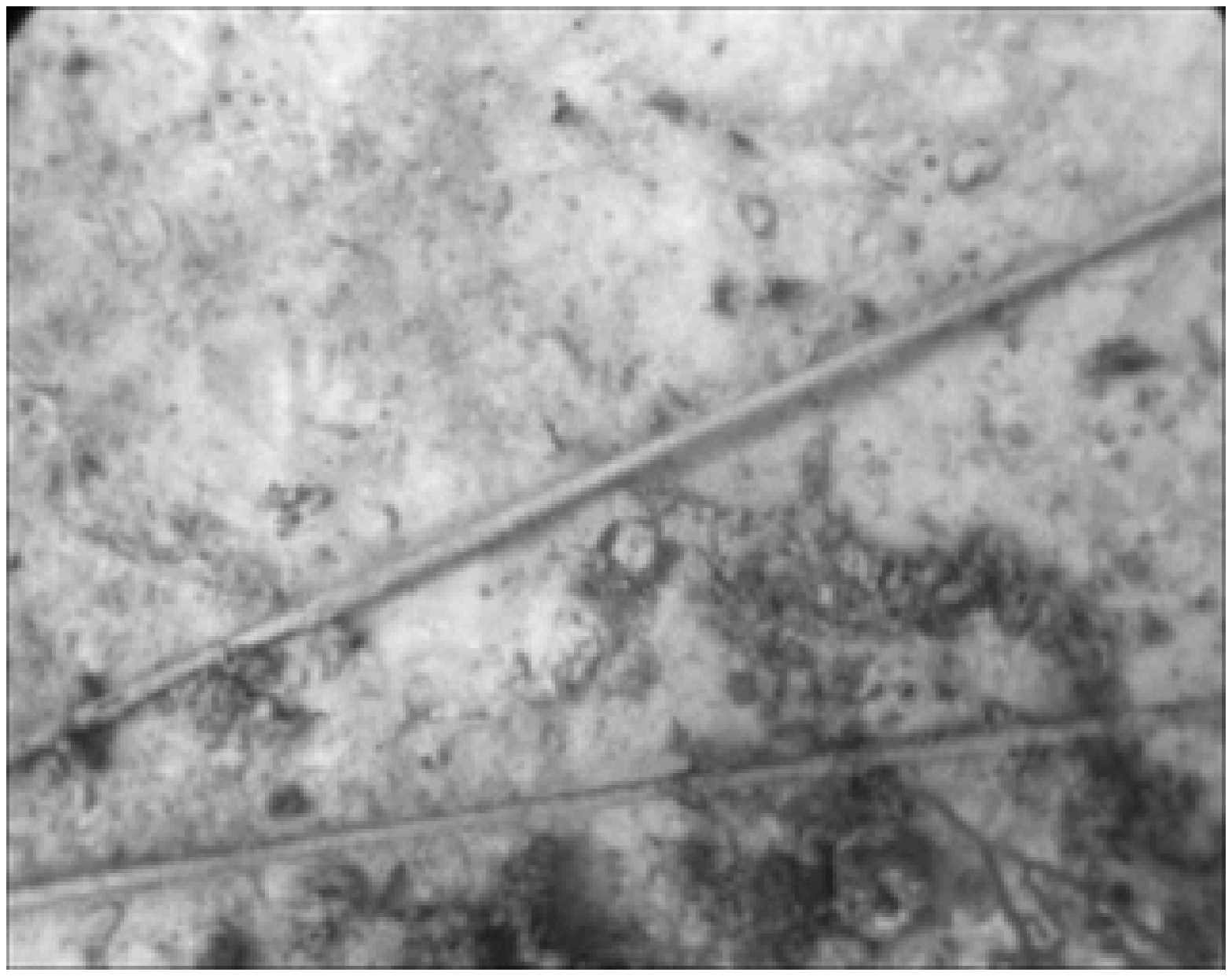,width=4cm}
\epsfig{figure=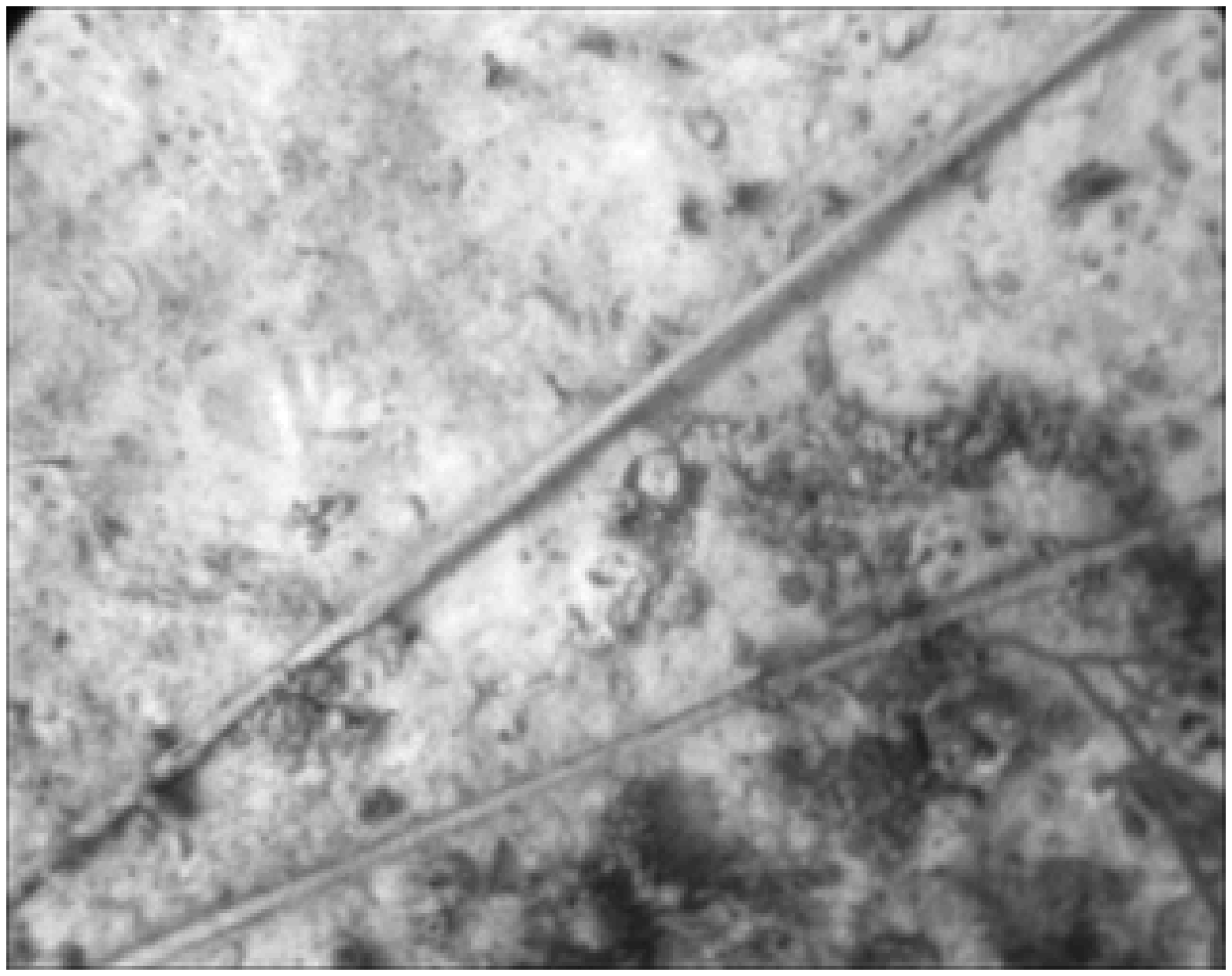,width=4cm}
\epsfig{figure=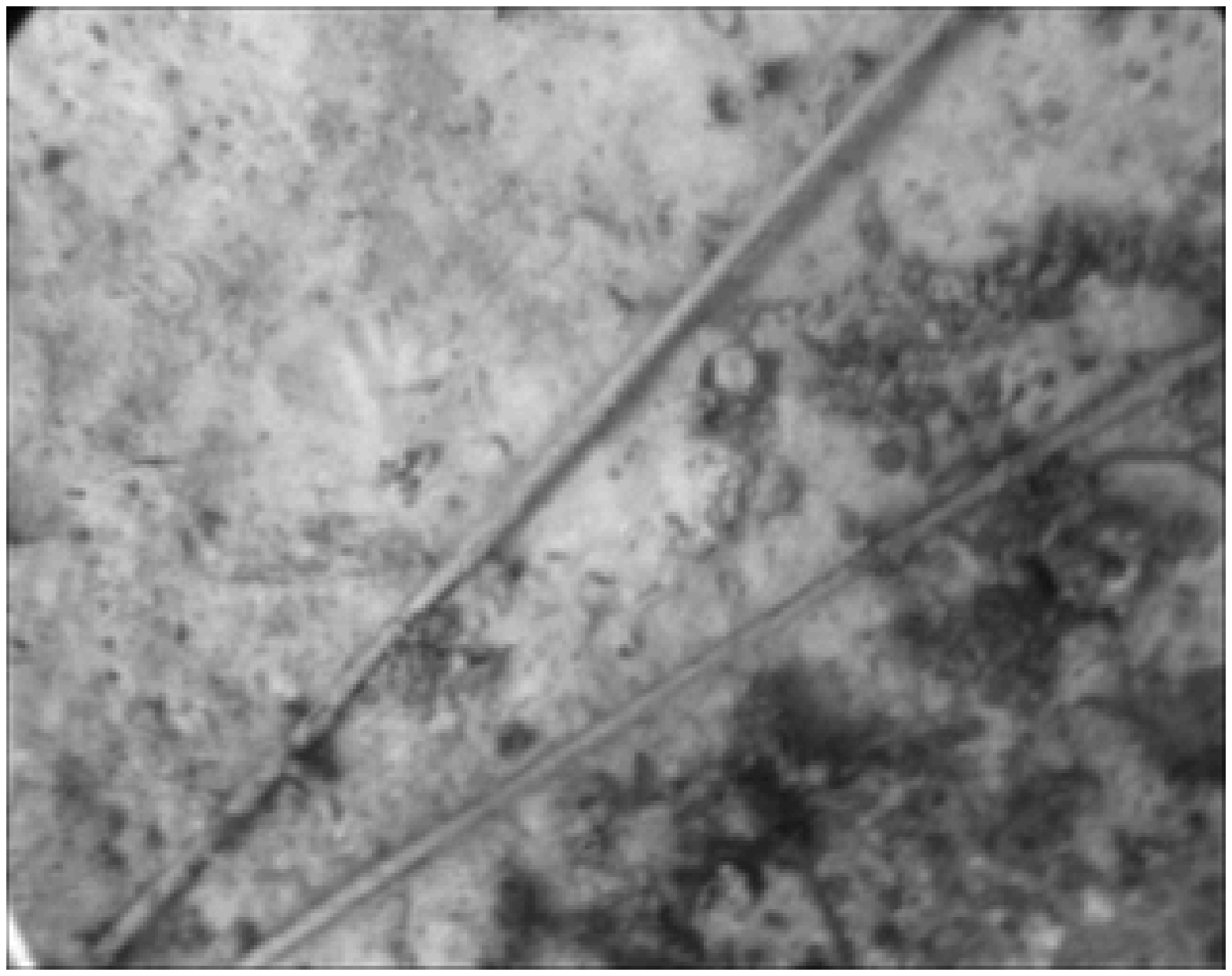,width=4cm}} \caption{Sample underwater
 images.} \label{fig:mosaic}
\end{figure}

\begin{figure}[hbt]
%\centerline{\epsfig{figure=under_12.eps,width=5cm}}
%\centerline{\epsfig{figure=under_123.eps,width=5cm}}
\centerline{\epsfig{figure=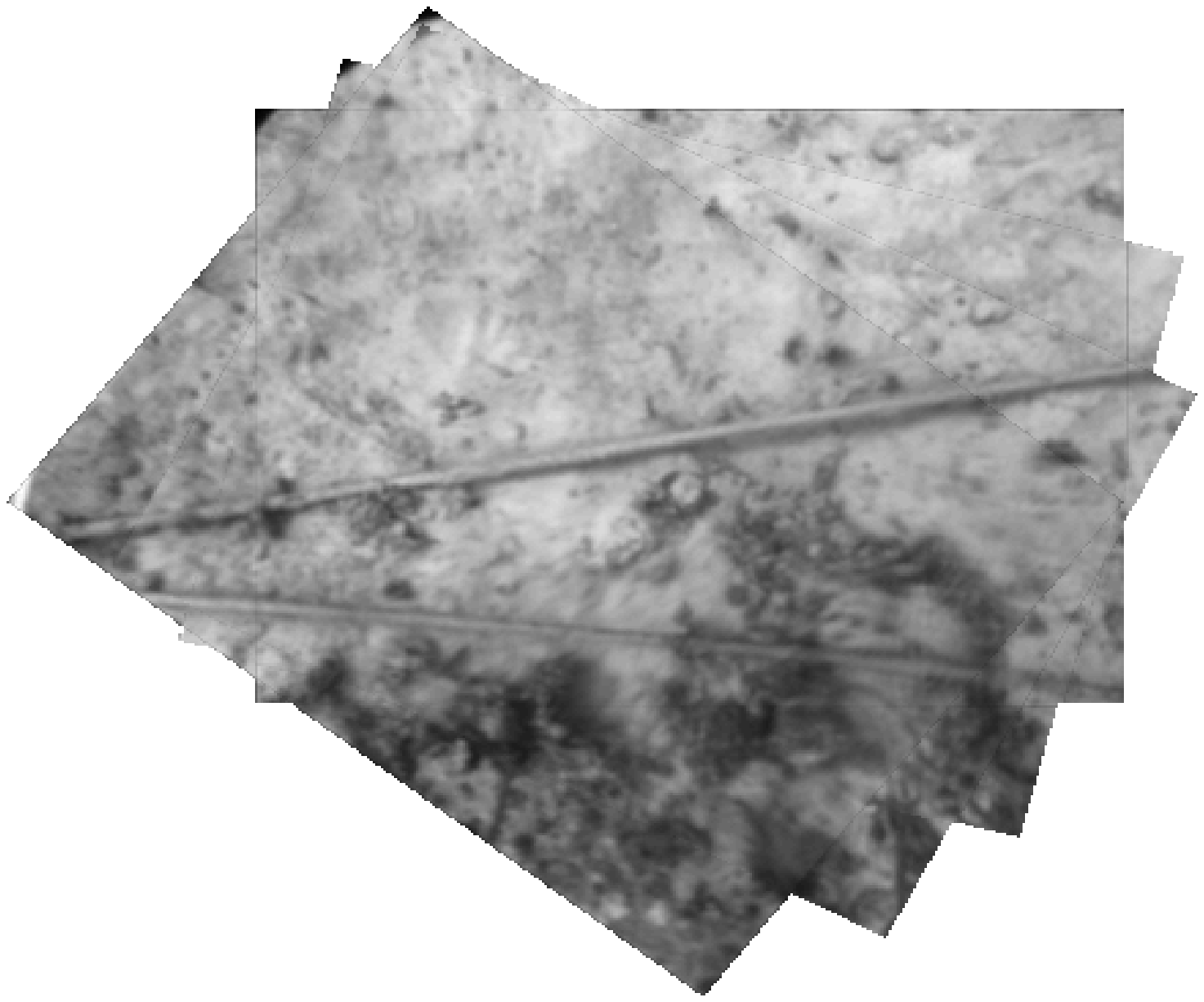,width=12cm}}
\caption{Mosaic built by using our algorithm to register the
images of Fig.~\ref{fig:mosaic}.} \label{fig:mosaic2}
\end{figure}

As a final example, we use images captured by an underwater camera
in the sea. Fig.~\ref{fig:mosaic} shows four of those images. The
low texture and the almost total absence of salient feature points
make these images particularly challenging. Note also that
although the overlapping region between images is not very small,
its shape is not rectangular. In this situation, the traditional
fixed window method would use a small rectangular window inside
the overlapping region, thus failing to use all the information
available.

In Fig.~\ref{fig:mosaic2}, we represent the seabed mosaic obtained
by using our algorithm to sequentially register the images of
Fig.~\ref{fig:mosaic}.

\section{Conclusion}
\label{sec:conc}

We proposed a new method to build a panoramic image from a set of
partial views. Rather than composing the input images in an
incremental way, our approach seeks the global solution to the
estimation problem, {\it i.e.}, it computes the panorama that best
matches all the partial observations. To minimize the global cost,
we derived an efficient gradient descent algorithm that
generalizes the current most robust two-frame featureless
registration approaches.

\bibliography{mosaics}

\begin{thebibliography}{10}
\providecommand{\url}[1]{#1}
\csname url@rmstyle\endcsname
\providecommand{\newblock}{\relax}
\providecommand{\bibinfo}[2]{#2}
\providecommand\BIBentrySTDinterwordspacing{\spaceskip=0pt\relax}
\providecommand\BIBentryALTinterwordstretchfactor{4}
\providecommand\BIBentryALTinterwordspacing{\spaceskip=\fontdimen2\font plus
\BIBentryALTinterwordstretchfactor\fontdimen3\font minus
  \fontdimen4\font\relax}
\providecommand\BIBforeignlanguage[2]{{%
\expandafter\ifx\csname l@#1\endcsname\relax
\typeout{** WARNING: IEEEtran.bst: No hyphenation pattern has been}%
\typeout{** loaded for the language `#1'. Using the pattern for}%
\typeout{** the default language instead.}%
\else
\language=\csname l@#1\endcsname
\fi
#2}}

\bibitem{aguiarjasinschimourapluem04}
P.~Aguiar, R.~Jasinschi, J.~Moura, and C.~Pluempitiwiriyawej, ``Content-based
  image sequence representation,'' in \emph{Digital Video Processing}, T.~Reed,
  Ed.\hskip 1em plus 0.5em minus 0.4em\relax CRC Press, 2004.

\bibitem{aguiar97}
P.~Aguiar and J.~Moura, ``Detecting and solving template ambiguities in motion
  segmentation,'' in \emph{IEEE ICIP}, 1997.

\bibitem{aguiar01-icip}
------, ``Image motion estimation -- convergence and error analysis,'' in
  \emph{IEEE ICIP}, Greece, 2001.

\bibitem{bergen92}
J.~Bergen, P.~Anandan, K.~Hanna, and R.~Hingorani, ``Hierarchical model-based
  motion estimation,'' in \emph{European Conf. on Computer Vision}, Santa
  Margherita Ligure, Italy, 1992.

\bibitem{dufaux00}
F.~Dufaux and J.~Konrad, ``Efficient, robust, and fast global motion estimation
  for video coding,'' \emph{IEEE T-IP}, 2000.

\bibitem{hartley2000}
R.~Hartley and A.~Zisserman, \emph{Multiple View Geometry in Computer
  Vision}.\hskip 1em plus 0.5em minus 0.4em\relax Cambridge University Press,
  2000.

\bibitem{hasler99}
D.~Hasler, L.~Sbaiz, S.~Ayer, and M.~Vetterli, ``From local to global
  pparameter estimation in panoramic photographic reconstruction,'' in
  \emph{IEEE ICIP}, Kobe, Japan, 1999.

\bibitem{jojic01}
N.~Jojic and B.~Frey, ``Learning flexible sprites in video layers,'' in
  \emph{IEEE Int.~Conf.~on CVPR}, Hawaii, 2001.

\bibitem{kim03}
D.~Kim and K.~Hong, ``Fast global registration for image mosaicing,'' in
  \emph{IEEE ICIP}, Barcelona, Spain, 2003.

\bibitem{lee02}
J.~Lee and J.~Ra, ``Block motion estimation based on selective integral
  projections,'' in \emph{IEEE ICIP}, Rochester, USA, 2002.

\bibitem{mann97}
S.~Mann and R.~Piccard, ``Video orbits of the projective group: a simple
  approach to featureless estimation of parameters,'' \emph{IEEE Trans.~on
  Image Processing}, 1997.

\bibitem{petrovic04}
N.~Petrovic, N.~Jojic, and T.~Huang, ``Hierarchical video clustering,'' in
  \emph{IEEE MMSP}, Siena, Italy, 2004.

\bibitem{pires04}
B.~Pires and P.~Aguiar, ``Registration of images with small overlap,'' in
  \emph{Proc. of the IEEE Multimedia Signal Processing Workshop}, Siena, Italy,
  2004.

\bibitem{pires05}
------, ``Featureless global alignment of multiple images,'' January 2005,
  submitted to {\it IEEE Int. Conf. on Image Processing}.

\bibitem{reddy96}
B.~Reddy and B.~Chattery, ``An {FFT}-based technique for translation, rotation,
  and scale-invariant image registration,'' \emph{IEEE Trans.~on Image
  Processing}, 1996.

\bibitem{shi94}
J.~Shi and C.~Tomasi, ``Good features to track,'' in \emph{IEEE Int. Conf. on
  Computer Vision and Pattern Recognition}, 1994.

\end{thebibliography}
\bibliographystyle{IEEEtranS}

\end{document}